\pdfoutput=1
\documentclass[11pt,breaklines,colorlinks,citecolor=blue]{article}

\usepackage[]{acl}
\usepackage{times,todonotes}
\usepackage{latexsym,multirow,booktabs}
\usepackage{subcaption}

\usepackage{wrapfig}
\usepackage{xspace}
\usepackage{xcolor,colortbl}
\definecolor{mygray}{gray}{0.85}

\usepackage{url}
\usepackage{todonotes}
\usepackage{float}

\usepackage[T1]{fontenc}
\usepackage[utf8]{inputenc}

\usepackage{microtype}
\usepackage{comment}
\usepackage{arydshln}
\newcommand{\name}{{\mbox{WALNUT}}\xspace}

\definecolor{forestgreen}{rgb}{0.13, 0.55, 0.13}

\newcommand{\gz}[1]{\textcolor{orange}{\bf\small [GZ: #1]}}

\newcommand{\update}[1]{\textcolor{black}{ #1}}
\newcommand{\labelfunsmall}[1]{{\scriptsize \textsf{#1}}}

\newcommand{\nerlabelto}[1]{{\scriptsize \textsf{#1}}}

\hypersetup{
        pdftitle={WALNUT: A Benchmark on Semi-weakly Supervised Learning for Natural Language Understanding},
        pdfauthor={Guoqing Zheng, Giannis Karamanolakis, Kai Shu, Ahmed Hassan Awadallah},
        pdfsubject={Proceedings of the 2022 Annual Conference of the North American Chapter of the Association for Computational Linguistics}
}

\title{\name: A Benchmark on Semi-weakly Supervised Learning for Natural Language Understanding}

\author{Guoqing Zheng \\
  Microsoft Research\\
  \texttt{zheng@microsoft.com} \\\And
  Giannis Karamanolakis \\
  Columbia University\\
  \texttt{gkaraman@cs.columbia.edu} \\\AND
  Kai Shu \\
  Illinois Institute of Technology \\
  \texttt{kshu@iit.edu} \\\And
  Ahmed Hassan Awadallah \\
  Microsoft Research \\
  \texttt{hassanam@microsoft.com} \\  
}

\begin{document}

\maketitle
\begin{abstract}

Building machine learning models for natural language understanding
(NLU) tasks relies heavily on labeled data. Weak supervision has been
proven valuable when large amount of labeled data is unavailable or
expensive to obtain. 
Existing works studying weak supervision for NLU
either mostly focus on a specific task or simulate weak supervision signals from ground-truth labels. 
It is thus hard to compare different approaches and evaluate the benefit of weak supervision without access to a unified and systematic benchmark with diverse tasks and real-world weak labeling rules.
In this paper, we propose such a benchmark,
named \name\footnote{\name: Semi-\textbf{W}e\textbf{A}kly
supervised \textbf{L}earning for \textbf{N}atural
language \textbf{U}nderstanding \textbf{T}estbed}, to advocate and
facilitate research on weak supervision for NLU. \name consists of NLU
tasks with different types, including document-level and
token-level prediction tasks. 
\name is the first semi-weakly supervised learning
benchmark for NLU, where each task contains weak labels generated
by multiple real-world weak sources, together with a small set of
clean labels. 
We conduct baseline evaluations on \name to
systematically evaluate the effectiveness of various weak supervision methods and model architectures. 
Our results demonstrate the benefit of weak supervision for low-resource NLU tasks and highlight interesting patterns across tasks.
We expect \name to stimulate further research on methodologies to
leverage weak supervision more effectively.
The benchmark and code for baselines are available at \url{aka.ms/walnut_benchmark}.

\end{abstract}

\section{Introduction}

To tackle natural language understanding (NLU) tasks via supervised learning, high-quality labeled
examples are crucial. Recent advances on large pre-trained language models~\citep{peters2018deep,devlin2018bert,radford2019language} lead to impressive gains on NLU benchmarks,
including GLUE~\citep{wang2018glue} and
SuperGLUE~\citep{wang2019superglue}, at the assumption that large
amount of labeled examples are available. 
For many real-world applications, however, it is expensive and time-consuming to manually obtain large-scale high-quality labels, while it is relatively easier to obtain auxiliary supervision signals, or \textit{weak
  supervision}, as a viable source to boost model performance without expensive data annotation process.

\begin{figure}[t]%
    \centering
    \includegraphics[width=\linewidth]{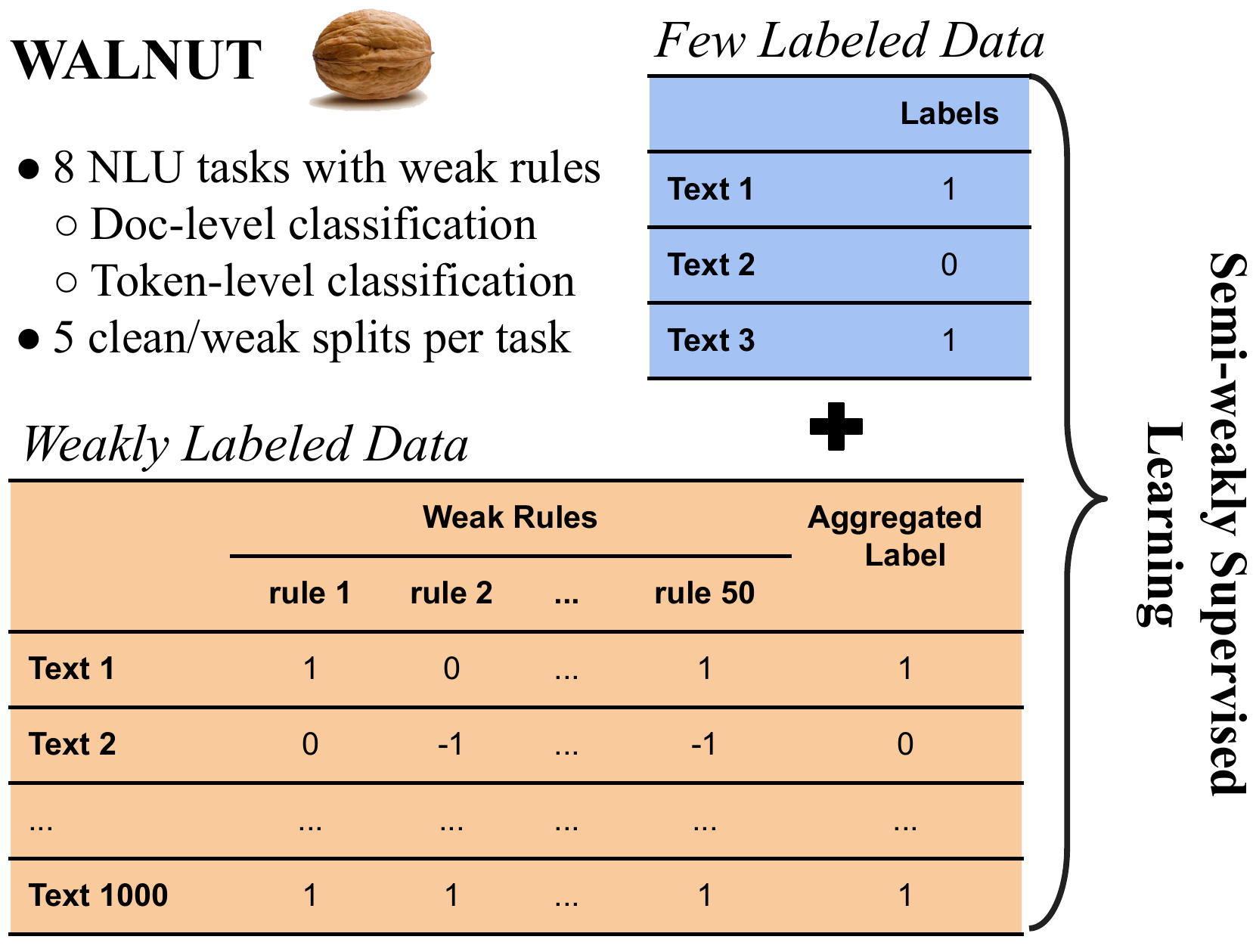}
    \caption{\name, a benchmark with 8 NLU tasks with real-world weak labeling rules. Each task in \name comes with few labeled data and weakly labeled data for semi-weakly supervised learning.} 
    \label{fig:walnut}
\end{figure}

Learning from weak supervision for NLU tasks is attracting increasing attention. Various types of weak supervision have been considered, such as knowledge bases~\citep{mintz2009distant,xu2013filling}, keywords~\citep{karamanolakis2019leveraging,ren2020denoising}, regular expression patterns~\citep{augenstein2016stance}, and other metadata such as user interactions in social media~\citep{shu2017fake}. 
\update{Also, inspired by recent advances from semi-supervised learning, \textit{semi-weakly supervised learning} methods which leverage both a small set of clean labels and a larger set of weak supervision}~\cite{papandreou2015weakly, hendrycks2018using,shu2019meta,pmlr-v139-mazzetto21a,karamanolakis2021self,maheshwari-etal-2021-semi,zheng2021mlc} are emerging to further boost task performance.
However, a unified and systematic evaluation benchmark supporting \textit{both
weakly and semi-weakly} supervised learning for NLU tasks is rather
limited.  On the one hand, many existing works only study specific NLU
tasks with weak supervision, thus evaluations of proposed techniques
leveraging weak supervision on a small set of tasks do not necessarily
generalized onto other NLU tasks.
On the other hand, some works rely on simulated weak supervision, such as weak labels
corrupted from ground-truth labels~\citep{hendrycks2018using}, while
real-world weak supervision signals can be far more complex than
simulated ones. Furthermore, existing weakly and semi-weakly supervised approaches
 are evaluated on different data with different metrics
and weak supervision sources, making it difficult to understand and compare.

To better advocate and facilitate research on leveraging weak
supervision for NLU, in this paper we propose \name (Figure \ref{fig:walnut}), a semi-weakly
supervised learning benchmark of NLU tasks with real-world weak
supervision signals. 
Following the tradition of existing benchmarks
(e.g., GLUE), we propose to cover different types of NLU tasks and domains, including document-level classification tasks (e.g.,
sentiment analysis on online reviews, fake news detection on news
articles), and token-level classification tasks (e.g., named entity
recognition in news and biomedical documents).
\name provides few labeled and many weakly labeled examples (Figure~\ref{fig:walnut}) and encourages a consistent and robust evaluation of different techniques, as we will describe in Section~\ref{sec:walnut}.  

In addition to the proposed benchmark, in Section~\ref{s:eval-results} we shed light on the benefit of weak supervision for NLU tasks in a collective manner, by evaluating several representative weak and semi-weak supervision methods for and several base models of various sizes (e.g., BiLSTM, BERT, RoBERTa), leading to more than 2,000 groups of experiments. 
Our large-scale analysis demonstrates that weak supervision is valuable for low-resource NLU tasks and that there is large room for performance improvement, thus motivating future research. 
Also, by computing the average performance across tasks and model architectures, we show surprising new findings. 
First, simple techniques for aggregating multiple weak labels (such as unweighted majority voting) achieve better performance than more complex weak supervision paradigms. 
Second, weak supervision has smaller benefit in larger base models such as RoBERTa, because larger pre-trained models can already achieve impressively high performance using just a few clean labeled data and no weakly labeled data at all.
We identify several more challenges on leveraging weak supervision for NLU
tasks and shed light on possible
future work based on \name.

The main contributions of this paper are: (1) We propose a new benchmark on semi-weakly supervised learning for NLU, which covers eight established annotated datasets and various text genres, dataset sizes, and degrees of task difficulty; (2) We conduct an exploratory analysis from different perspectives to demonstrate and analyze the results for several major existing weak supervision approaches across tasks; and (3) We discuss the benefits and provide insights for potential weak supervision studies for representative NLU tasks.

\section{Related Work}

\subsection{Weak Supervision for NLU}

\paragraph{Document-level classification} 

Existing works on weakly supervised learning for document-level classification attempt to correct the weak labels by incorporating a loss correction mechanism for text classification~\cite{sukhbaatar2014training,patrini2017making}.  Other works further assume access to a small set of clean labeled examples~\cite{hendrycks2018using,ren2018learning,varma2018snuba,shu2020learning}. 
Recent works also consider the scenario where weak signals are available from multiple sources~\cite{ratner2017snorkel,meng2018weakly,ren2020denoising}.
Despite the recent progress on weak supervision for text classification, there is no agreed upon benchmark that can guide future directions and development of NLU tasks in semi-weakly supervised setting.

\paragraph{Token-level classification}

Weak supervision has also been studied for token-level classification (sequence tagging) tasks, focusing on Named Entity Recognition (NER).
\update{
One of the most common approaches is distant
supervision~\citep{mintz2009distant}, which uses knowledge bases to heuristically annotate training data.
Besides distant supervision, several weak supervision approaches have recently addressed NER by introducing various types of labeling rules, for example based on keywords, lexicons, and regular expressions~\citep{fries2017swellshark,ratner2017snorkel, shang2018automated,safranchik2020weakly,lison2020named,li-etal-2021-bertifying}.}
\name integrates existing weak rules into a unified representation and evaluation format.

\subsection{NLU Benchmarks}

Accompanying the emerging of large pre-trained language models, NLU
benchmarks has been a focus for NLP research, including
GLUE~\cite{wang2018glue} and
SuperGLUE~\cite{wang2019superglue}. On such benchmarks, the major focus
is put on obtaining best possible performance~\cite{he2020deberta} under the full
training setting, which assumes that a large quantity of manually labeled
examples are available for all tasks.
Few-shot NLU benchmarks exist~\cite{FewGLUE, FewCLUE, ye2021crossfit, CLUES}, however these do not contain weak supervision.
Though research in weak
supervision in NLU has gained significant interest~\cite{hendrycks2018using,shu2019meta,zheng2021mlc}, most of these work
either focus on a small set of tasks or simulate weak supervision
signals from ground-truth labels, hindering its generalization ability
to real-world NLU tasks. The lack of a unified test bed covering
different NLU task types and data domains motivates us to construct
such a benchmark to better understand and leverage semi-weakly supervised learning
for NLU in this paper.

Different from existing work based on
crowd-sourcing~\citep{hovy2013learning,
gokhale2014corleone} to obtain noisy labels, we focus specifically on the \textit{semi-weakly supervised}
learning setting, where we collect tasks with weak labels obtained from human-written labeling rules. 
~\cite{zhang2021wrench} is concurrent work that also features weak supervision for various (not necessarily text-based) tasks and assumes a purely weakly supervised setting, i.e., no clean labeled data is available. In contrast, \name focuses on NLU tasks under a more-realistic semi-weakly supervised setting and, as we show in Section~\ref{sec:walnut}, a small amount of clean labeled data plays an important role in determining the benefit of weak supervision for a target task.

\section{\name}
\label{sec:walnut}

\begin{table*}[tbp!]
\centering \small
\caption{Statistics of the eight document- and token-level tasks in \name. See Section~\ref{s:task-categories} for details.}
\resizebox{\linewidth}{!}{
\begin{tabular}{lcccccccc}
\toprule
Dataset & AGNews &IMDB & Yelp  & GossipCop & CoNLL& NCBI& WikiGold& LaptopReview\\
\midrule
 Label granularity & doc.& doc.&   doc.&  doc.& token& token& token& token\\
 Task & topic& sentiment& sentiment  &fake  & NER& NER& NER& NER\\
 Domain & news& movies& restaurants  &news  & news& biomed& web& tech\\
 \# Classes & 4& 2& 2  &2  & 9& 3& 9& 3\\
  \# Train-clean ($|D_C|$)& 80& 40& 40  & 40 & 180& 60& 360 & 150\\
 \# Train-weak ($|D_W|$) & 4,439& 16,626& 10,954& 6,462 &  13,861& 532 & 995 & 2,286\\
 \# Dev  & 12,000& 2,500 & 3,800& 1,430 & 3,250 & 99& 169&  609\\
 \# Test  & 12,000& 2,500 & 3,800& 957& 3,453 & 99 & 170& 800\\
 \# Weak rules & 9&8 &8& 3 &50 &12 & 16& 12\\
\bottomrule
\end{tabular}}
\label{tab:doc_data}
\end{table*}

\subsection{Benchmark Construction Principles}
We first describe the design principles guiding the benchmark construction.
\paragraph{Task Selection Criterion} We aim to create a testbed which covers a broad range of NLU tasks where \emph{real-world weak supervision signals are available}. To this end, 
\name includes eight English text understanding tasks from diverse domains, ranging from news articles, movie reviews, merchandise reviews, biomedical corpus, wikipedia documents, to tweets. The eight tasks are categorized evenly into two types, namely document classification and token classification (sequence labeling). It's worth noting that \textit{we didn't create any labeling rules ourselves to avoid bias, but rather opted with labeling rules which already exist and are extensively studied by previous research.}
\update{Therefore, \name does not include other NLU tasks, such as natural language inference and question answering, as we are not aware of previous research with human labeling rules for these tasks.}

\paragraph{Semi-weakly Supervised Learning Setting} While many previous works studied weak supervision in a purely weakly supervised setting, recent advances in few-shot and semi-supervised learning suggest that a small set of cleanly labeled examples together with unlabeled examples greatly helps boosting the task performance. Though large scale labeled examples for a task is difficult to collect, we acknowledge that it's rather practical to collect a small set of labeled examples. In addition, recent methods leveraging weak supervision also demonstrate greater gains of combining a small set of labeled examples with large weakly labeled examples~\cite{hendrycks2018using,shu2019meta,zheng2021mlc}. Therefore, \name is designed to emphasize the semi-weakly supervised learning setting. Specifically, each dataset contains both a small
 number of clean labeled instances and a large number of weakly-labeled
 instances. Each weakly-labeled instance comes with multiple weak labels
 (assigned by multiple rules) and a single aggregated weak label derived from weak rules. \textit{Note that this way \name can be naturally used to support the conventional weakly supervised setting by ignoring the provided clean labels.}

 \paragraph{Consistent and Robust Evaluation} To address discrepancies in evaluation protocols from existing research on weak supervision and to better account for the small set of clean examples per task, \name is constructed to promote systematic and robust evaluations across all eight tasks. Specifically, for each task, we first determine the number of clean examples to sample with pilot experiments (with the rest treated as weakly labeled examples by applying the corresponding weak labeling rules), such that the weakly supervised examples can be still helpful with the small clean examples present (typically 20-50 per class; see Sec. \ref{s:dataset-preprocessing} for details); second, to consider sampling uncertainty, we repeat the sampling process for the desired number of clean examples 5 times and provide all 5 splits in \name. Methods on \name are expected to be using all 5 pre-computed splits and reporting the mean and variance of its performance.

To summarize, \name can facilitate research on weakly- and semi-weakly supervised learning by offering the following:
 \begin{itemize}
   \item Eight NLU tasks from diverse domains;
   \item For each task, five pairs of clean and weakly labeled samples for robust evaluation;
   \item For each individual weakly labeled example, all weak labels from multiple rules and a single aggregated weak label.
 \end{itemize}

\subsection{Task Categories}
\label{s:task-categories}
Here, we describe the eight tasks in \name (Table~\ref{tab:doc_data}), grouped into four document-level classification tasks (Section~\ref{s:document-level-tasks}) and four token-level classification tasks (Section~\ref{s:task-token-level}).

\subsubsection{Document-level Classification}
\label{s:document-level-tasks}
The goal of document-level classification tasks is to classify a sequence of tokens $x_1, \dots, x_N$ to a class $c \in C$, where $C$ is a pre-defined set of classes. 
We consider binary and multi-class classification problems from different application domains such as sentiment classification~\citep{zhang2015character}, fake news detection~\citep{shu2020leveraging}, and topic classification~\citep{zhang2015character}. 
Concretely, we include the following widely-used document-level text classification datasets: AGNews~\cite{zhang2015character}, Yelp~\cite{zhang2015character}, IMDB~\cite{maas2011learning} and GossipCop~\cite{shu2020fakenewsnet}.

For Yelp, IMDB, and AGNews, the weak rules are derived from the text using keyword-based heuristics, third-party tools as detailed in~\citep{ren2020denoising}.
For GossipCop, the weak labeling rules are derived from social context information accompanying the news articles, including related users' social engagements on the news items (e.g., user comments in Twitter).  For example, a weak labeling rule for fake news can be ``If a news piece has a standard deviation of user sentiment scores greater than a threshold, then the news is weakly labeled as fake news. ''~\citep{shu2020leveraging}. 

\subsubsection{Token-level Classification}
\label{s:task-token-level}
The goal of token-level classification tasks is to classify a sequence of tokens $x_1, \dots, x_N$ to a sequence of tags $y_1, \dots, y_N \in C'$, where $C'$ is a pre-defined set of tag classes (e.g., person or organization). 
As one of the most common token-level classification tasks, Named Entity Recognition (NER) deals with recognizing categories of named entities (e.g., person, organization, location) and is important in several NLP pipelines, including information extraction and question answering.

We include in \name the following four NER datasets from different domains, for which weak rules are available: CoNLL~\cite{sang2003introduction}, the NCBI Disease corpus~\cite{dougan2014ncbi}, WikiGold~\cite{balasuriya2009named} and the LaptopReview corpus~\cite{pontiki2016semeval} from the SemEval 2014 Challenge.
For the CoNLL and WikiGold dataset, we use weak rules provided by~\cite{lison2020named}. 
For the NCBI and LaptopReview dataset, we use weak rules provided by~\cite{safranchik2020weakly}.

\begin{figure*}[t]
    \centering
        \begin{subfigure}[t]{0.245\linewidth}
        \includegraphics[width=\linewidth]{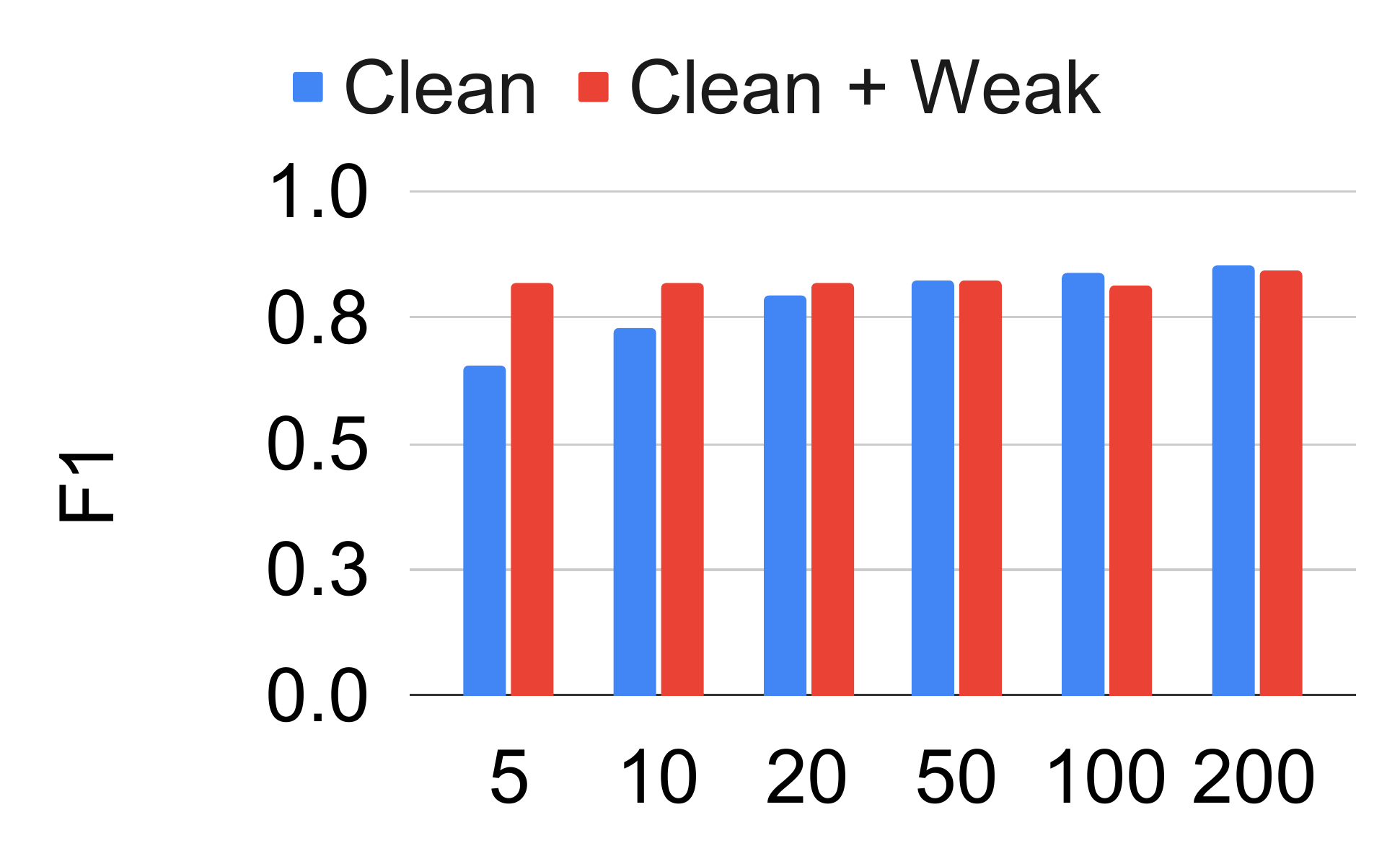}
        \caption{AG News}
        \label{fig:train-size-AG}
        \end{subfigure}
        \begin{subfigure}[t]{0.245\linewidth}
        \includegraphics[width=\linewidth]{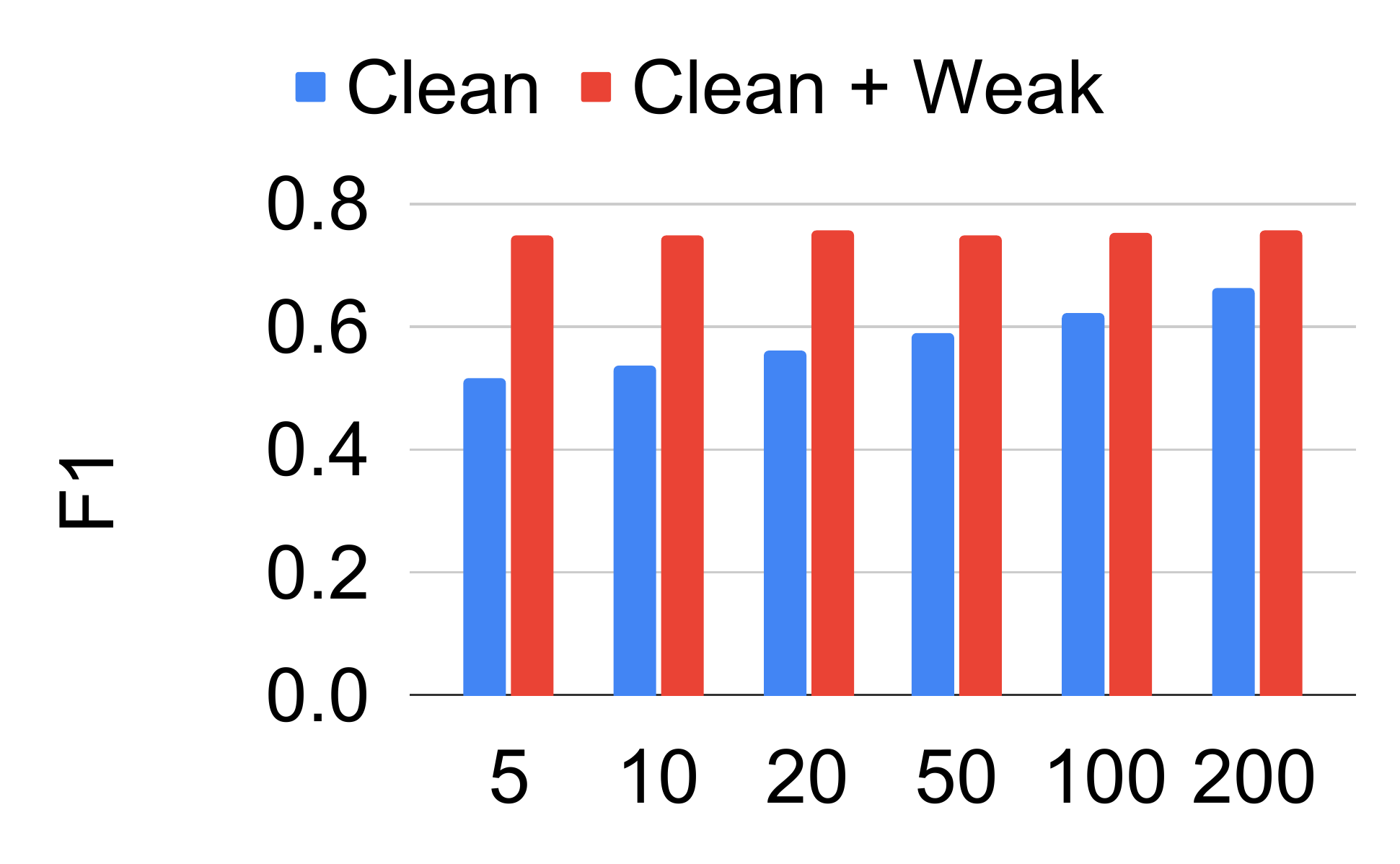}
        \caption{IMDB}
        \label{fig:train-size-IMDB}
        \end{subfigure}
        \begin{subfigure}[t]{0.235\linewidth}
        \includegraphics[width=\linewidth]{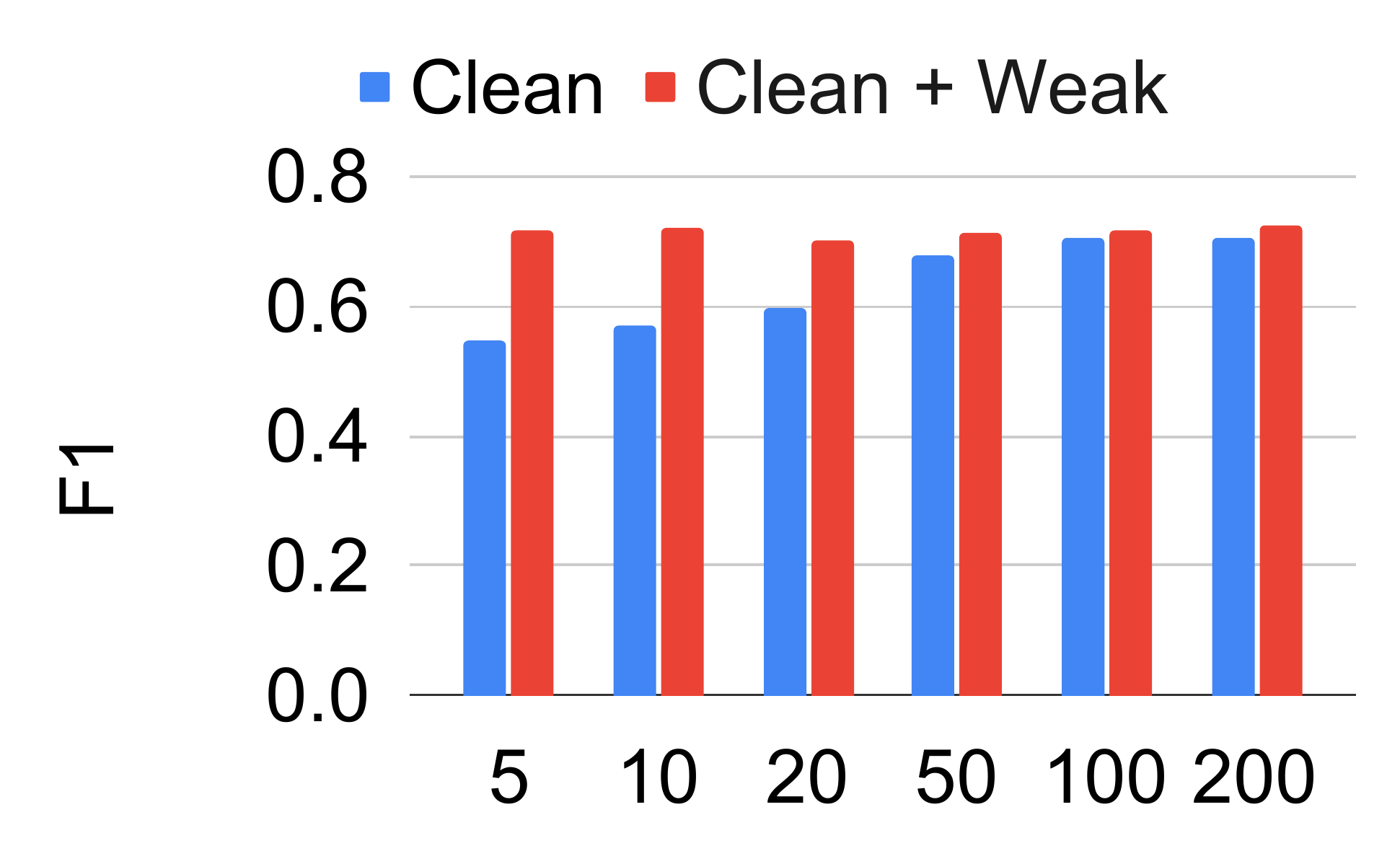}
        \caption{Yelp}
        \label{fig:train-size-yelp}
        \end{subfigure}
        \begin{subfigure}[t]{0.235\linewidth}
        \includegraphics[width=\linewidth]{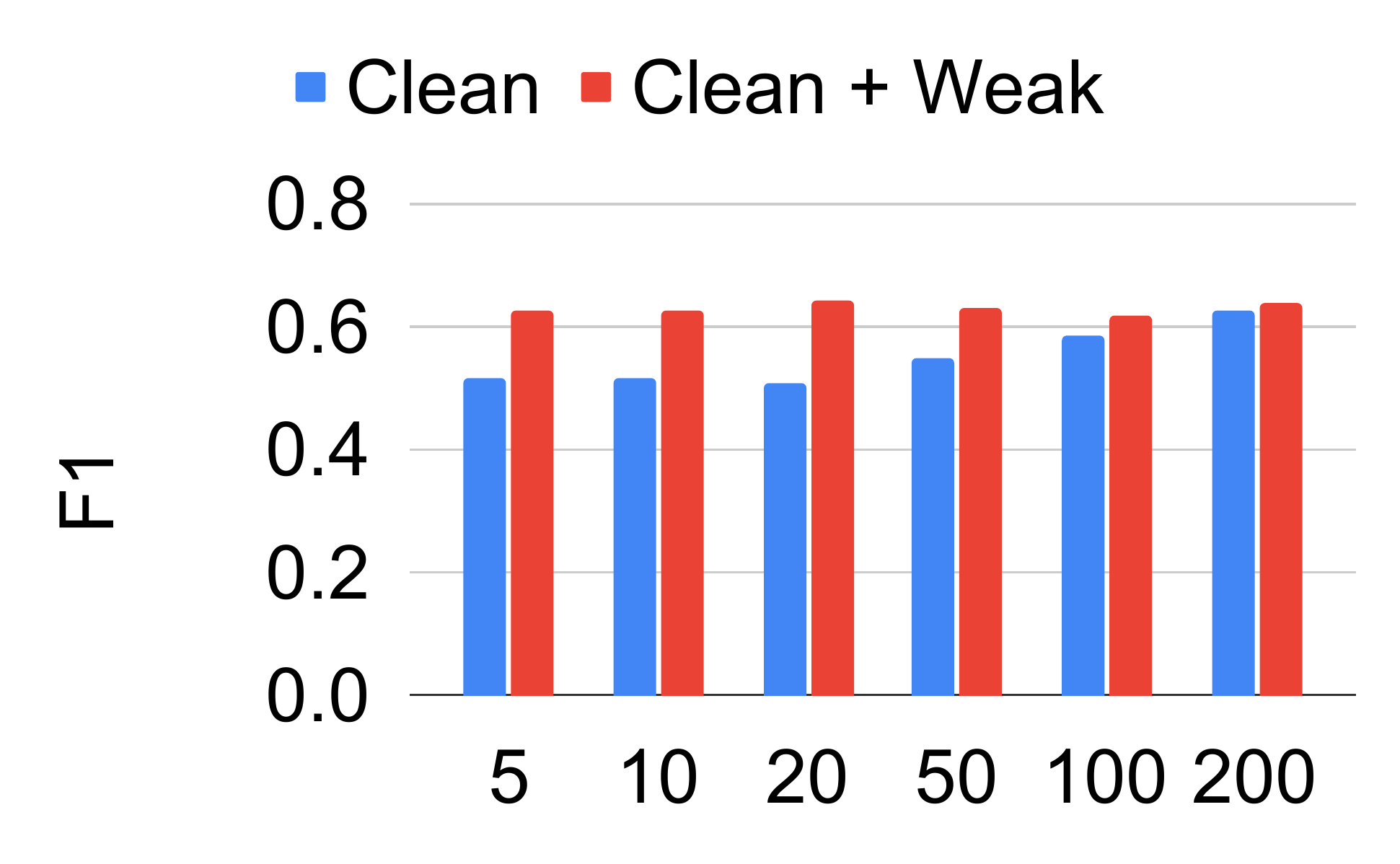}
        \caption{Gossipcop}
        \label{fig:train-size-gossipcop}
        \end{subfigure}
        \begin{subfigure}[t]{0.245\linewidth}
        \includegraphics[width=\linewidth]{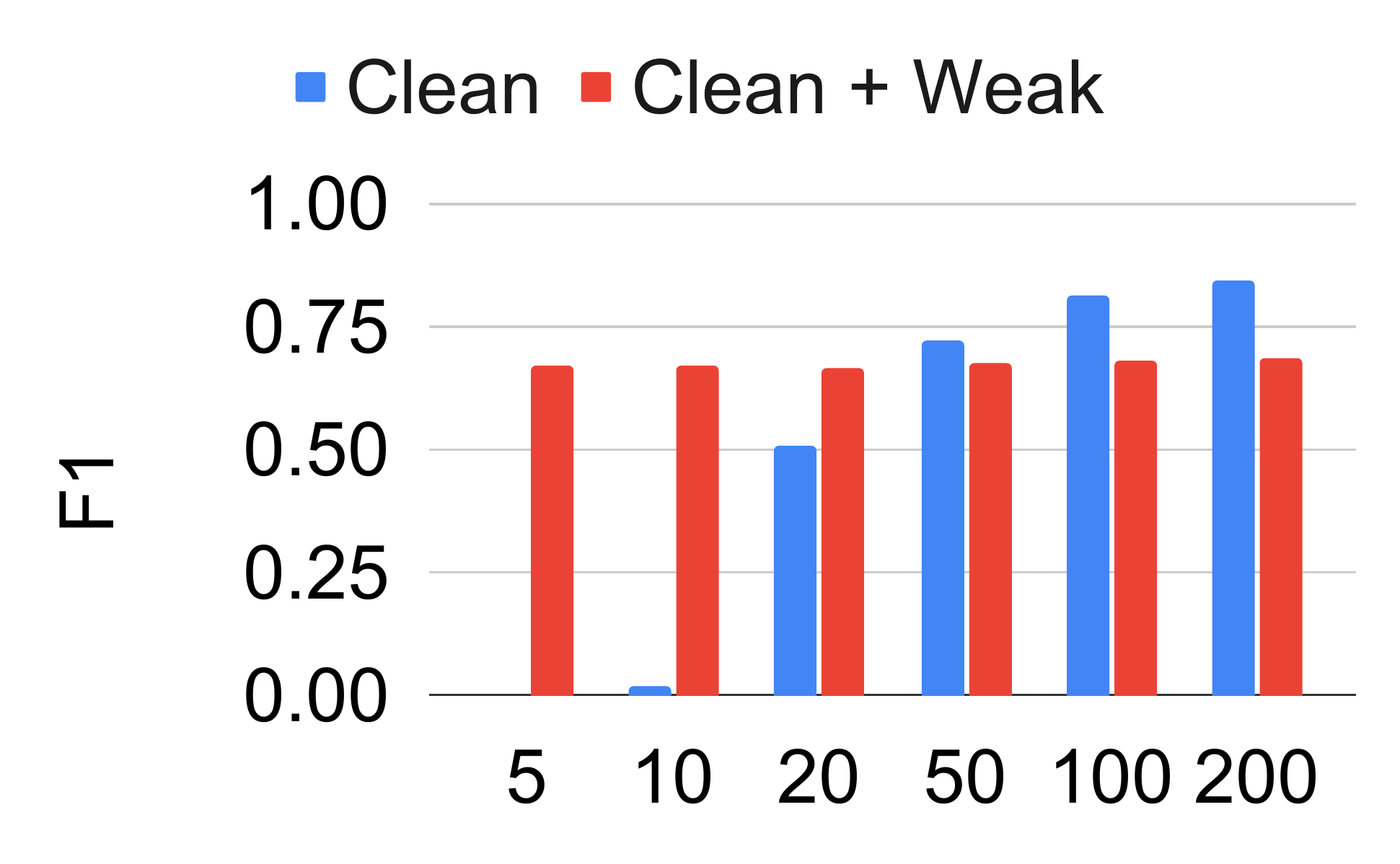}
        \caption{CoNLL}
        \label{fig:train-size-CoNLL}
        \end{subfigure}
        \begin{subfigure}[t]{0.245\linewidth}
        \includegraphics[width=\linewidth]{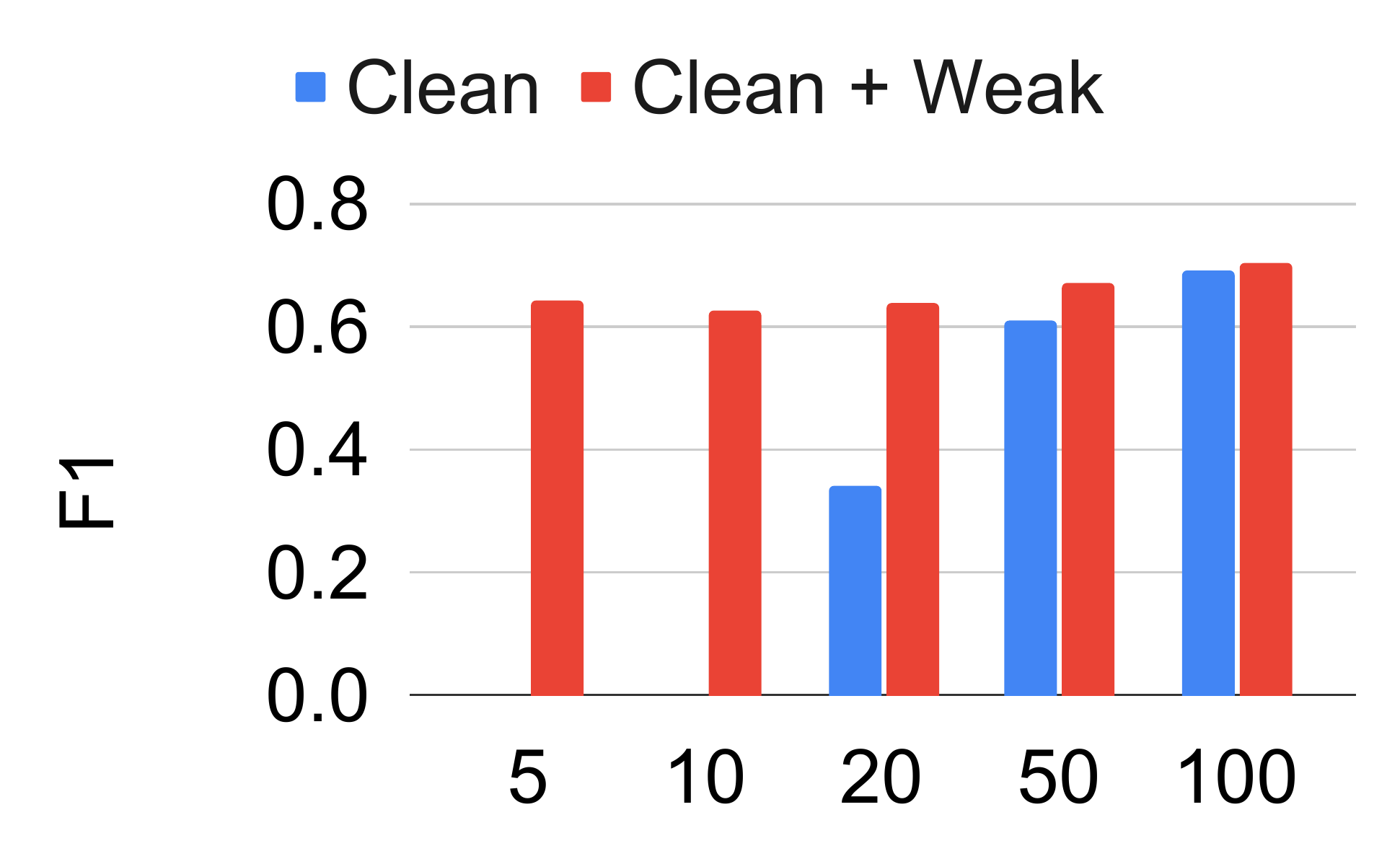}
        \caption{NCBI}
        \label{fig:train-size-NCBI}
        \end{subfigure}
        \begin{subfigure}[t]{0.245\linewidth}
        \includegraphics[width=\linewidth]{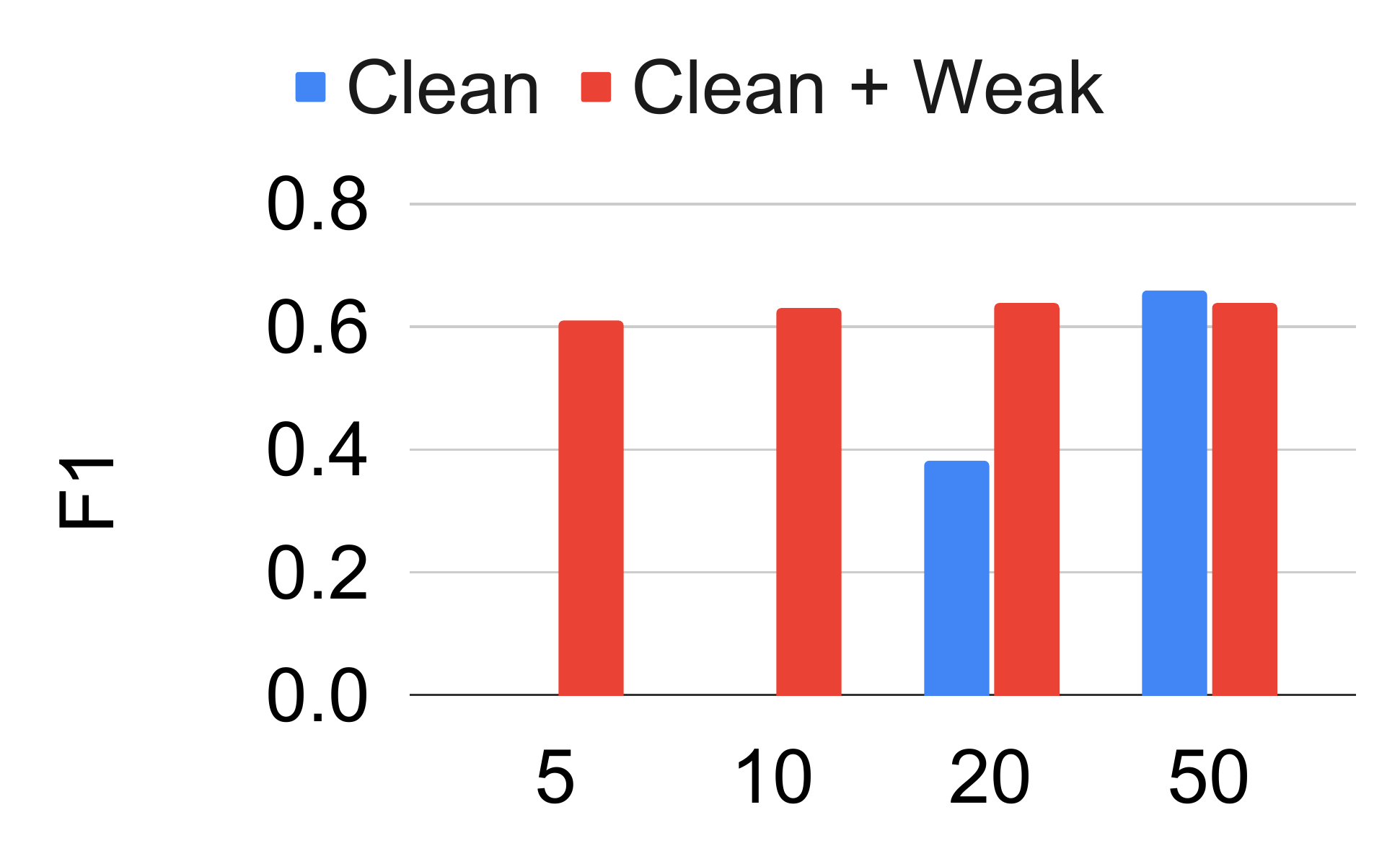}
        \caption{WikiGold}
        \label{fig:train-size-wikigold}
        \end{subfigure}
        \begin{subfigure}[t]{0.245\linewidth}
        \includegraphics[width=\linewidth]{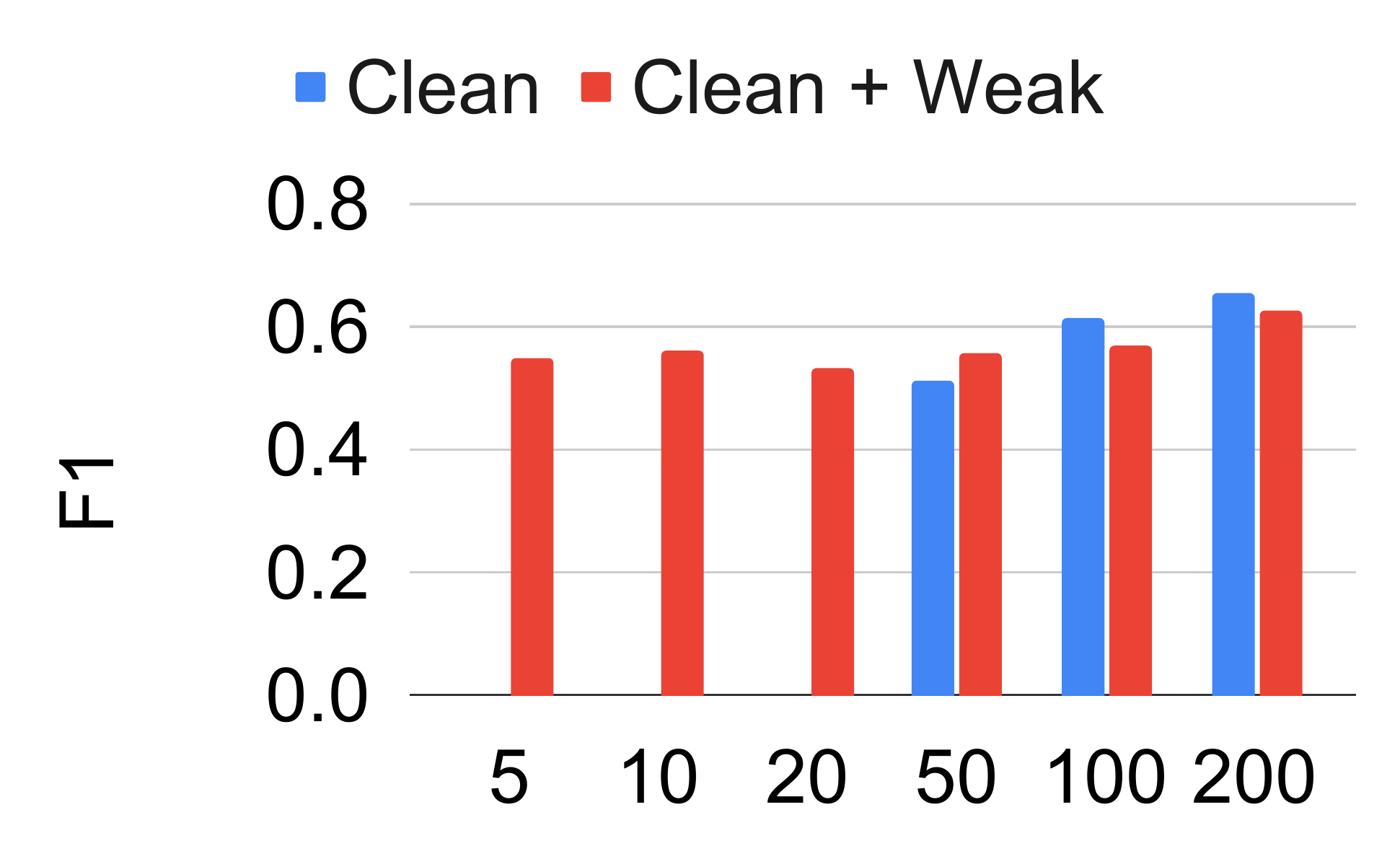}
        \caption{LaptopReview}
         \label{fig:train-size-LaptopReview}
        \end{subfigure}
    \caption{F1 score by varying (in the x-axis) the number of clean instances per class considered in the clean training set ($D_C$). The importance of weak supervision is more evident for settings with smaller numbers of instances, where the gap in performance between the ``Clean'' approach and ``Clean+Weak'' approach is larger. For a robust evaluation across tasks, \name provides five clean/weak splits per task. See Section~\ref{s:dataset-preprocessing} for details.} 
    \label{fig:all_cw}
\end{figure*}

\subsection{Dataset Pre-Processing}
\label{s:dataset-preprocessing}

To construct a semi-weakly supervised learning setting, we split the training dataset for each task into a small subset with clean labels ($D_C$) and a large subset with weak labels ($D_W$). 
For robust evaluation, we create five different clean/weak train splits as we noticed that the model performances may vary with different clean train instances. The validation/test sets are always the same across splits. 

Because of different dataset characteristics (e.g., differences in number of classes, difficulty),  we choose the size for $D_C$ per dataset via pilot studies. (After having selected the instances for the $D_C$, we consider the remaining instances as part of the $D_W$ split.) 
We defined the size of $D_C$ such that we demonstrate the benefits of weak supervision and at the same time leave substantial room for improvement in future research. To this end, we compare the performances of the same base classification model (e.g., BiLSTM), trained using only $D_C$ (``Clean'' approach) v.s. using both $D_C$ and $D_W$ (``Clean+Weak'' approach). As shown in Figure~\ref{fig:all_cw}, for each dataset, we choose a small size of $D_C$, such that the ``Clean+Weak'' approach has a substantially higher F1 score than the ``Clean'' approach and at the same time the ``Clean'' approach has no trivial F1 score.

The statistics of the pre-processed datasets included in \name are shown in Table~\ref{tab:doc_data}.

\section{Baseline Evaluation in \name}
In this section, we describe the baselines and evaluation procedure (Section~\ref{s:baselines}), and discuss evaluation results in \name (Section~\ref{s:eval-results}). Our results highlight the value of weak supervision, important differences across different baselines, and the potential utility of \name for future research on weak supervision. 

\subsection{Baselines and Evaluation Procedure}
\label{s:baselines}
We evaluate several baseline approaches in \name by considering different base models (text encoders) and different (semi-)weakly supervised learning methods to train the base model. 

\paragraph{Encoder Models}
To encode input text, we experiment with various text encoders, ranging from shallow LSTMs to large pre-trained
transformer-based encoders~\citep{vaswani2017attention}.  In particular, we consider
a series of models with increasing model size: Bi-directional LSTM with Glove embeddings~\cite{pennington2014glove}, DistilBERT~\cite{sanh2019distilbert}, BERT~\cite{devlin2018bert}, RoBERTa~\cite{liu2019roberta}, BERT-large, and RoBERTa-large. For each text encoder, a classification head is placed on top of the encoder to perform the task. For more details on the base model configurations see Appendix \ref{s:imple_detail}.

\paragraph{Learning Methods}
Given the semi-weakly supervised setting in \name, we evaluate eight supervision approaches in the following categories:
\begin{itemize}
\item \textit{Learning from clean labeled examples only}. The model is trained on only the small amount of available clean examples $D_C$, a naive baseline method leveraging no weak supervision, which we denote as \textbf{C}.
\item \textit{Learning from weakly labeled examples only}. The model is trained on all weakly labeled examples $D_W$. To produce a single weak label from the multiple labeling rules for training, we aggregate the rules via two methods: majority voting (denoted by \textbf{W}) and Snorkel~\cite{ratner2017snorkel} (denoted by \textbf{Snorkel}).
\item \textit{Learning from both clean and weakly labeled examples}. The model is trained with both $D_C$ and $D_W$ in a weakly-supervised setting. The first two baselines in this category is simply concatenating $D_C$ and the aggregated weak labels (from either \textbf{W} and \textbf{Snorkel}), and the model is trained on the combination. We denote these two as \textbf{C+W} and \textbf{C+Snorkel}, respectively. We also test three recent semi-weakly supervised learning methods which proposed better ways to leverage both $D_C$ and $D_W$: \textbf{GLC} which is a loss correction approach~\cite{hendrycks2018using}, \textbf{MetaWN} which is a meta-learning approach to learn the importance of weakly labeled examples~\cite{shu2019meta, ren2018learning} and \textbf{MLC}, a meta-learning approach to learn to correct the weak labels~\cite{zheng2021mlc}. 
\end{itemize}
To establish an estimate of the ceiling performance on \name, for each task we also train with all clean training examples in the original dataset (denoted by \textbf{Full Clean}).
\paragraph{Experimental Procedure}
For a robust evaluation, we repeat each experiment five times on the
five splits of $D_C$ and $D_W$ (clean and weak examples for each task; see
Section~\ref{s:dataset-preprocessing}), and report the average scores and the standard deviation across the five runs.  In \name, we report the average micro-average F1 score on the test set.\footnote{For
token-level F1, we use the conlleval implementation: \url{https://huggingface.co/metrics/seqeval}} Datasets and code for \name are publicly available at \url{aka.ms/walnut_benchmark}.

\begin{table*}[!htb]%
  \small
  \caption{Main results on \name with F1 score (in $\%$) on all tasks. The rightmost column reports the average F1 score across all tasks. (MLC is not shown for BERT-large and RoBERTa-large due to OOM.)}
    \label{tab:walnut-main-results}
\resizebox{\linewidth}{!}{
\begin{tabular}{r|cccccccc|c}
\toprule
    Method & AGNews & IMDB & Yelp  & GossipCop & CoNLL & NCBI & WikiGold & LaptopReview & AVG \\\midrule
    \multicolumn{10}{c}{BiLSTM (20M parameters)} \\
    \rowcolor{mygray}
Full Clean & 89.4 & 83.1 & 86.4 & 64.5      & 31.9  & 69.9 &   21.8 & 62.6         & 63.7 \\
C          & 79.5 & 56.2 & 59.5 & 50.8      & 00.8  & 58.2 &    \textbf{15.8}& 42.3         &  45.4\\
W          & 78.0 & 75.2 & 70.8 & 62.0      & 11.1  & 52.3 &    02.7& 49.4         & 50.2 \\
Snorkel       & 79.9 & 75.4 & \textbf{76.0} & 61.4      & 06.7  & 52.5 & 02.7  & 49.4         & 52.1 \\
C+W        & 82.0 & \textbf{75.6} & 70.2 & \textbf{64.1}      & \textbf{17.2}  & 56.8 &  15.7  & 51.2         & \textbf{52.5} \\
C+Snorkel     & \textbf{82.9} & 75.4 & 66.5 & 62.6      & 07.7  & \textbf{59.2} &   10.7 & 53.8         &  52.4\\
GLC        & 56.5 & 72.2 & 63.7 & 60.5      & 05.1  & 58.9 &   08.7 & \textbf{55.2}         & 47.6 \\
MetaWN     & 55.2 & 72.7 & 65.5 & 58.2      & 00.0  & 53.9 & 03.4  & 51.6         &  45.1\\
MLC        & 55.3 & 72.3 & 65.7 & 52.5      & 00.0  & 52.5 &  05.9  & 51.5         & 43.7 \\\hline

\multicolumn{10}{c}{DistilBERT-base (66M parameters)} \\
    \rowcolor{mygray}
Full Clean & 92.1 & 88.8 & 93.7 & 75.1      & 88.6  & 75.7 &  79.7  & 75.8         & 83.7 \\

C          & 80.8 & 71.2 & 73.1 & 55.3      & 51.4  & 57.7 &  69.5  & 53.0         & 64.0 \\
W          & 72.2 & 75.0 & 70.2 & 70.8      & 66.9  & 62.0 &  57.4  & 53.8         & 66.0 \\
Snorkel       & 70.2 & 70.7 & 65.9 & 68.4      & 64.3  & 62.9 &  56.3  & 54.0         & 65.1 \\
C+W        & 83.3 & 74.8 & 71.5 & \textbf{71.4}      & 66.9  & 66.2 &  64.0  & 57.3         & 68.5 \\
C+Snorkel     & \textbf{84.3} & \textbf{81.7} & \textbf{81.8} & 69.1      & 64.6  & 67.8 &  64.4 & 57.5         & \textbf{71.4}\\
GLC        & 67.8 & 74.1 & 68.1 & 67.3      & \textbf{72.4}  & \textbf{72.8} &  \textbf{71.7}  & \textbf{66.8}         &  70.1\\
MetaWN     & 70.0 & 74.4 & 69.3 & 70.0      & 65.7  & 64.2 & 58.5 & 58.2         & 66.3 \\
MLC        & 70.4 & 74.3 & 69.4 & 69.6      & 69.2  & 66.2 &  58.3  & 58.0         & 66.9\\\hline
 
\multicolumn{10}{c}{BERT-base (110M parameters)} \\
    \rowcolor{mygray}
Full Clean & 92.5 & 90.0   & 74.7 & 74.7      & 89.4  & 78.4 & 81.1  & 76.2         & 82.1 \\

C          & 82.9 & 63.8 & 60.3 & 57.1      & 67.3  & 66.6 & \textbf{71.9}  & 54.6         &  65.6\\
W          & 72.3 & 75.5 & 69.6 & 69.0      & 67.5  & 59.5 &  56.7 & 55.9         &  65.8\\
Snorkel       & 73.7 & 72.9 & 65.6 & 68.2      & 65.1  & 60.9 & 53.8  & 56.2         &  66.0\\
C+W        & \textbf{80.1} & 81.8 & 71.3 & 68.4      & 68.4  & 67.9 &  65.0 & 59.2         &  68.9\\
C+Snorkel     & 76.2 & \textbf{82.6} & \textbf{75.3} & 67.1      & 65.9  & 69.9 &   64.3 & 59.6         & 70.1 \\
GLC        & 68.8 & 75.7 & 68.8 & 68.1      & \textbf{74.7}  & \textbf{74.7} & 70.7 & \textbf{65.8}         & \textbf{70.9}\\
MetaWN     & 72.8 & 75.2 & 68.1 & 69.8      & 66.9  & 66.7 & 58.9 & 59.2         & 67.2 \\
MLC        & 73.0 & 74.7 & 70.0 & \textbf{71.3}      & 70.4  & 68.4 & 58.5 & 59.7         &  68.2\\\hline 

\multicolumn{10}{c}{RoBERTa-base (125M parameters)} \\
    \rowcolor{mygray}
Full Clean & 92.8 & 92.4 & 95.9 & 77.2      & 91.2  & 83.1 &  87.2 & 80.2         & 87.5 \\
C          & \textbf{84.1} & 74.5 & 70.2 & 57.4      & 72.9  & 72.9 &  78.2 & 61.3         & 71.4 \\
W          & 66.4 & 76.1 & \textbf{70.4} & 71.4      & 64.9  & 69.9 &  64.1  & 58.9         &  67.8\\
Snorkel       & 71.9 & 70.1 & 66.3 & 69.2      & 61.2  & 70.0 & 61.8  & 59.7         &  67.5\\
C+W        & 70.6 & \textbf{76.5} & \textbf{70.4} & 72.2      & 64.1  & 74.0 & 71.6  & 61.2         & 68.9 \\
C+Snorkel     & 74.6 & 68.2 & 66.4 & 71.4      & 62.2  & 73.4 &  72.2  & 61.6         & 68.8 \\
GLC        & 67.6 & 74.9 & 69.0 & 68.0      & \textbf{74.6}  & \textbf{79.1} &  \textbf{79.6}  & \textbf{71.5}         & \textbf{73.0}  \\
MetaWN     & 69.6 & 75.4 & 69.0 & 71.8      & 63.8  & 69.9 &  63.5  & 62.5         &  68.2\\
MLC        & 70.4 & 74.5 & 69.9 & \textbf{72.9}      & 68.3  & 74.3 & 63.1  & 63.6         &  69.6\\\hline

\multicolumn{10}{c}{BERT-large (336M parameters)} \\
\rowcolor{mygray}
Full Clean & 92.5&91.4 &94.9 &73.5 & 90.2&	80.5&	82.8&	78.9& 85.6	\\
C          & 72.5& 65.4& 68.4& 57.8& 67.2	 &69.7	 & \textbf{73.9}&	51.1 & 65.8\\
W          & 68.5& \textbf{75.9}& 70.7& 69.3& 65.7 & 62.0&57.1	&	54.2& 65.4 \\
Snorkel    & 73.3& 70.9& 65.8& 70.0& 63.6&	67.3&57.2	&	54.4 & 66.5\\
C+W        & 73.4& 74.8& \textbf{71.8}& 70.2& 66.7 &70.7&66.9	&	55.7& 67.6\\
C+Snorkel  & \textbf{73.6}& 71.3& 65.9& \textbf{71.3}& 63.6	&69.7	&63.4&	57.2& 67.0\\
GLC        & 67.1 & 74.6 & 67.3 & 69.8 & \textbf{71.8}	& \textbf{76.1}& 68.1	&	\textbf{65.4} & \textbf{70.0}\\
MetaWN    & 71.6 & 74.2 & 67.0 & 70.8 & 64.4 & 70.1	& 53.9&	45.9 & 64.7\\

\hline
\multicolumn{10}{c}{RoBERTa-large (355M parameters)} \\
    \rowcolor{mygray}
Full Clean & 93.1& 94.4& 96.9& 78.5& 91.3 &	83.5&	87.7&	80.4 & 88.2\\
C        & \textbf{86.1}& 69.1&\textbf{84.8} &69.1 & 76.4 &	77.7&	77.1&	60.6 &\textbf{75.1} \\
W         & 74.3& \textbf{77.7}& 70.5& 73.2& 63.6& 67.4 &  61.2&	57.3 & 68.2\\
Snorkel      & 75.5& 72.6& 67.1& 69.3& 61.1	&68.1 &	61.0 &	59.2 & 67.9 \\
C+W        & 71.9&77.4 &70.6 &71.6 & \textbf{63.8}& 71.2 &	70.4 &	60.2 & 68.5\\
C+Snorkel    & 74.0& 69.0& 66.5& \textbf{73.7}& 61.1	& 71.8 & 69.1 &	62.4 & 68.5\\
GLC        & 67.8 & 75.8 & 68.7 & 64.1 & 56.7 &	\textbf{80.0} &	\textbf{78.3} &	\textbf{68.4} & 70.0\\
MetaWN    & 68.6 & 66.2 & 71.1 & 64.6 & 63.2 & 69.5 &	59.3 &	61.5 & 65.5\\\hline 

\multicolumn{10}{c}{Rules (no base model)} \\\hline 
Rules  &  61.8& 73.9 & 65.9& 73.5&	61.3	& 	64.7	& 52.2 & 	60.0 & 64.1 \\
\bottomrule
\end{tabular}}
\end{table*}

\subsection{Experimental Results and Analysis}
\label{s:eval-results}

Table~\ref{tab:walnut-main-results} shows the main evaluation results on \name. Rows correspond to supervision methods for the base model, columns correspond to tasks, and each block corresponds to a different base model.
Unless explicitly mentioned, in the rest of this section we will compare approaches based on their average performance across tasks (rightmost column in Table~\ref{tab:walnut-main-results}).

As expected, training with Full Clean achieves the highest F1 score, corresponding to the high-resource setting where all clean labeled data are available. 
Such method is not directly comparable to the rest of the methods but serves as an estimate of the ceiling performance for \name.
Training with only limited clean examples achieves the lowest overall F1 score: in the low-resource setting, which is the main focus in \name, using just the available clean subset ($D_C$) is not effective.

\begin{table*}[htb!]%
\centering
\caption{Average F1 score across the eight tasks in \name. The bottom row computes the average F1 score across tasks and supervision methods. 
The three right-most columns report the average F1 score across model architectures and all tasks (``All''), document-level tasks (``Doc''), and token-level tasks (``Token'').
}
\label{tab:average-f1-std-results}

\resizebox{\linewidth}{!}{
\begin{tabular}{r|cccccc|ccc}
\toprule
 &\multicolumn{6}{c}{Base Model Architecture} & \multicolumn{3}{|c}{Average Results}\\
Method & BiLSTM & DistilBERT & BERT  & RoBERTa & BERT-large  & RoBERTa-large &  All  & Doc & Token\\
\midrule
\rowcolor{mygray}
Full Clean & 63.7 & 83.7 & 82.1 & 87.5 & 85.6 & 88.2 & 81.8 & 86.6 & 77.0 \\
C         & 45.4 & 64.0 & 65.6 & 71.4 & 65.8 & \textbf{75.1} & 64.5 & 68.7 & 60.3 \\
W         & 50.2 & 66.0 & 65.8 & 67.8 & 65.4 & 68.2 & 63.9 & 71.9 & 55.9 \\
Snorkel      & 52.1 & 65.1 & 66.0 & 67.5 & 66.5 & 67.9 & 64.2 & 70.4 & 57.9 \\
C+W       & \textbf{52.5} & 68.5 & 68.9 & 68.9 & 67.6 & 68.5 & 65.8 & \textbf{73.6} & 58.0 \\
C+Snorkel    & 52.4 & \textbf{71.4} & 70.1 & 68.8 & 67.0 & 68.5 & 66.3 & 73.0 & 59.7 \\
GLC       & 47.6 & 70.1 & \textbf{70.9} & \textbf{73.0} & \textbf{70.0} & 70.0 & \textbf{66.9} & 68.6 & \textbf{65.3} \\
MetaWN    & 45.1 & 66.3 & 67.2 & 68.2 & 64.7 & 65.5 & 62.8 & 69.2 & 56.4 \\
\midrule
AVG       & 51.1 &	69.4& 69.6 & 71.6 &69.1	 &71.5 & & & \\
\bottomrule
\end{tabular}}
\end{table*}

\begin{table*}[!htb]
\centering
\caption{Overall performance gain and gap of all weak supervision methods (Weak Sup, by averaging performance of W, Snorkel, C+W, C+Snorkel, GLC, MetaWN and MLC) against no weak supervision (C) and full clean training. Note that RoBERTa-large in included here, as the standard deviation of its performance with different splits on tasks varies significantly (See Table \ref{tab:std-results} in Appendix) hence using its performance mean as an indicator is less conclusive.}
\label{tab:benefit-of-weak-supervision}
\resizebox{\linewidth}{!}{
\begin{tabular}{l|cccccr}
\toprule
& BiLSTM & DistilBERT & BERT  & RoBERTa & BERT-large  & AVG  \\
\midrule 
Perf. gain: Weak Sup $-$ C & 3.90 &	4.21	& 2.98 &	-2.07 &	1.32&	 2.07\\
Perf. gap: Full Clean $-$ Weak Sup & 14.42	& 15.47 &	13.58	& 18.13&	18.51 & 16.00\\
\bottomrule
\end{tabular}}
\end{table*}
\paragraph{Weak supervision is valuable for low-resource NLU.}
``W'' and ``Snorkel'' achieve better F1 scores than ``C'' for many base models: even using only weakly-labeled data in $D_W$ is more effective than using just $D_C$, thus demonstrating that simple weak supervision approaches can be useful in the low-resource setting. 
Approaches such as ``C+W'' and ``C+Snorkel'' lead to further improvements compared to ``C'' and ``Snorkel'', thus highlighting that even simple approaches for integrating clean and weak labeled data (here by concatenating $D_C$ and $D_W$) are more effective than considering each separately. 

\paragraph{There is no clear winner in \name.}
Our results in Table~\ref{s:eval-results} indicate that the performance of weak supervision techniques varies substantially across tasks. 
Therefore, it is important to evaluate such techniques in a diverse set of tasks to achieve a fair comparison and more complete picture of their performance. 
The performance of various techniques also varies across different splits (See Table \ref{tab:std-results} in Appendix for variances of all experiments). 
Interestingly, ``C+W'' and ``C+Snorkel'' sometimes perform better than more complicated approaches, such as GLC, MetaWN and MLC.

\paragraph{Larger base models achieve better overall performance.}
We further aggregate statistics across tasks, methods, and base models in Table~\ref{tab:average-f1-std-results}.
The bottom row reports the average performance across methods for each base model and leads to a consistent ranking in F1 score among base models: BiLSTM $\leq$ DistilBERT $\leq$ BERT-base 
$\leq$ RoBERTa-base. 
Observing higher scores for larger transformer models such as RoBERTa agrees with previous observations~\citep{browngpt}. Interestingly, switching from BERT-base to BERT-large (and from RoBERTa-base to RoBERTa-large) in base model architecture leads to marginal improvement, suggesting the need to explore more effective learning methods leveraging weak supervision. 

\paragraph{Weak supervision has smaller benefit in larger base models.}
Another question that we attempt to address in \name is on whether weak supervision equally benefits each base model architecture. 
To quantify such benefit we compare the performance differences between models trained using semi-weak supervision and models trained using clean data only.
The ``Weak Sup'' approach in Table~\ref{tab:benefit-of-weak-supervision} is computed as the average F1 score across all semi-weak supervision methods (C+W, C+Snorkel, GLC, and MetaWN). 
The performance gap between ``Weak Sup'' and ``C'' (training with few clean data only) is smaller for larger models.
Additionally, the performance gap between ``Full Clean'' (full clean data training) and ``Weak Sup'' approach is larger for larger models. 
The two above observations highlight that weak supervision has smaller benefit in larger models.
An important future research direction is to develop better learning algorithms and improving the effectiveness of weak supervision in larger models. 
%
%
%
%
%

%
%
%

\begin{figure*}[!htb]
  \centering
        \begin{subfigure}[t]{0.495\linewidth}
        \includegraphics[width=\linewidth]{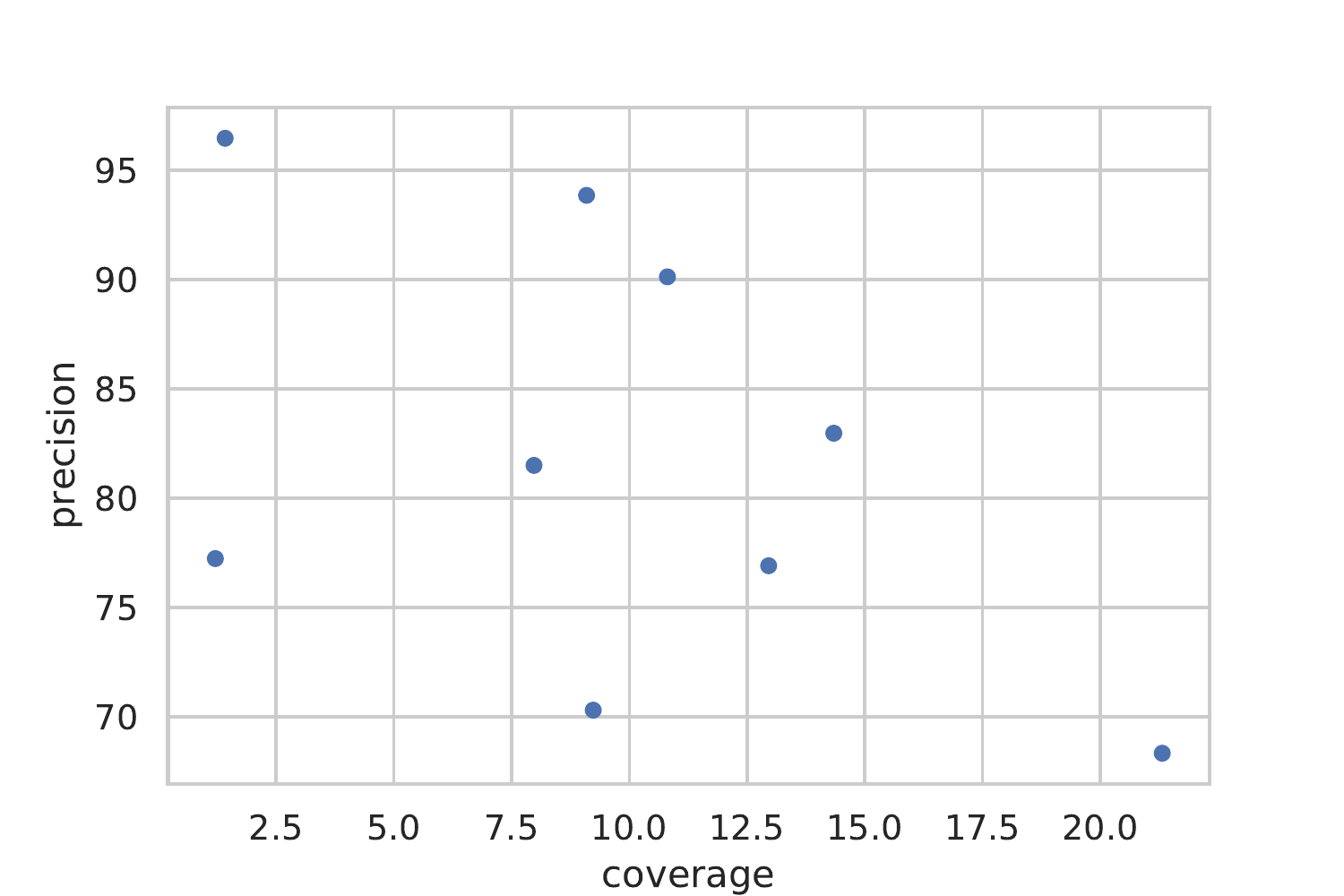}
        \caption{AG News.}
        \label{fig:rule-precision-recall-plot-agnews}
        \end{subfigure}
        \begin{subfigure}[t]{0.495\linewidth}
        \includegraphics[width=\linewidth]{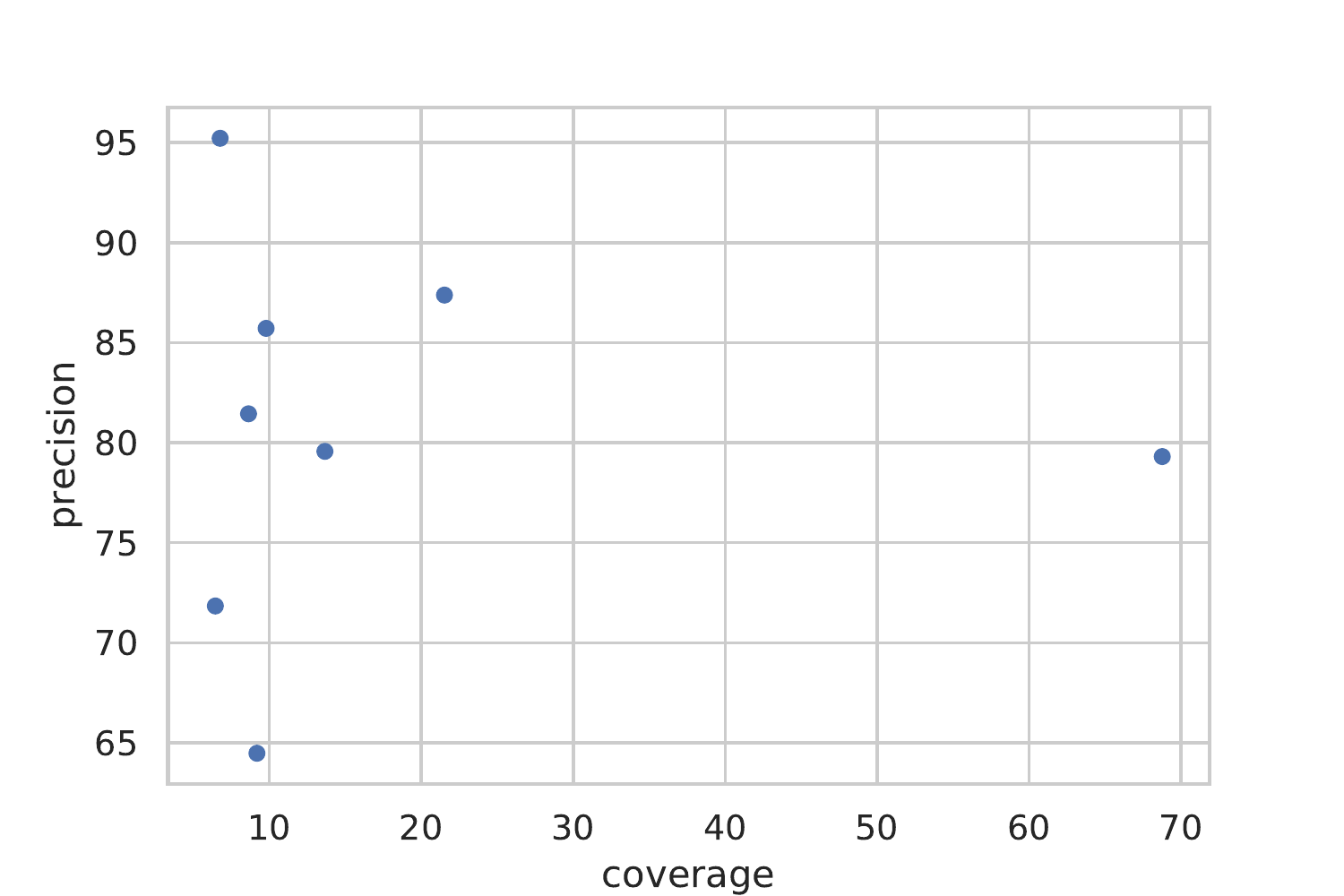}
        \caption{Yelp.}
        \label{fig:rule-precision-recall-plot-Yelp}
        \end{subfigure}
        \begin{subfigure}[t]{0.495\linewidth}
        \includegraphics[width=\linewidth]{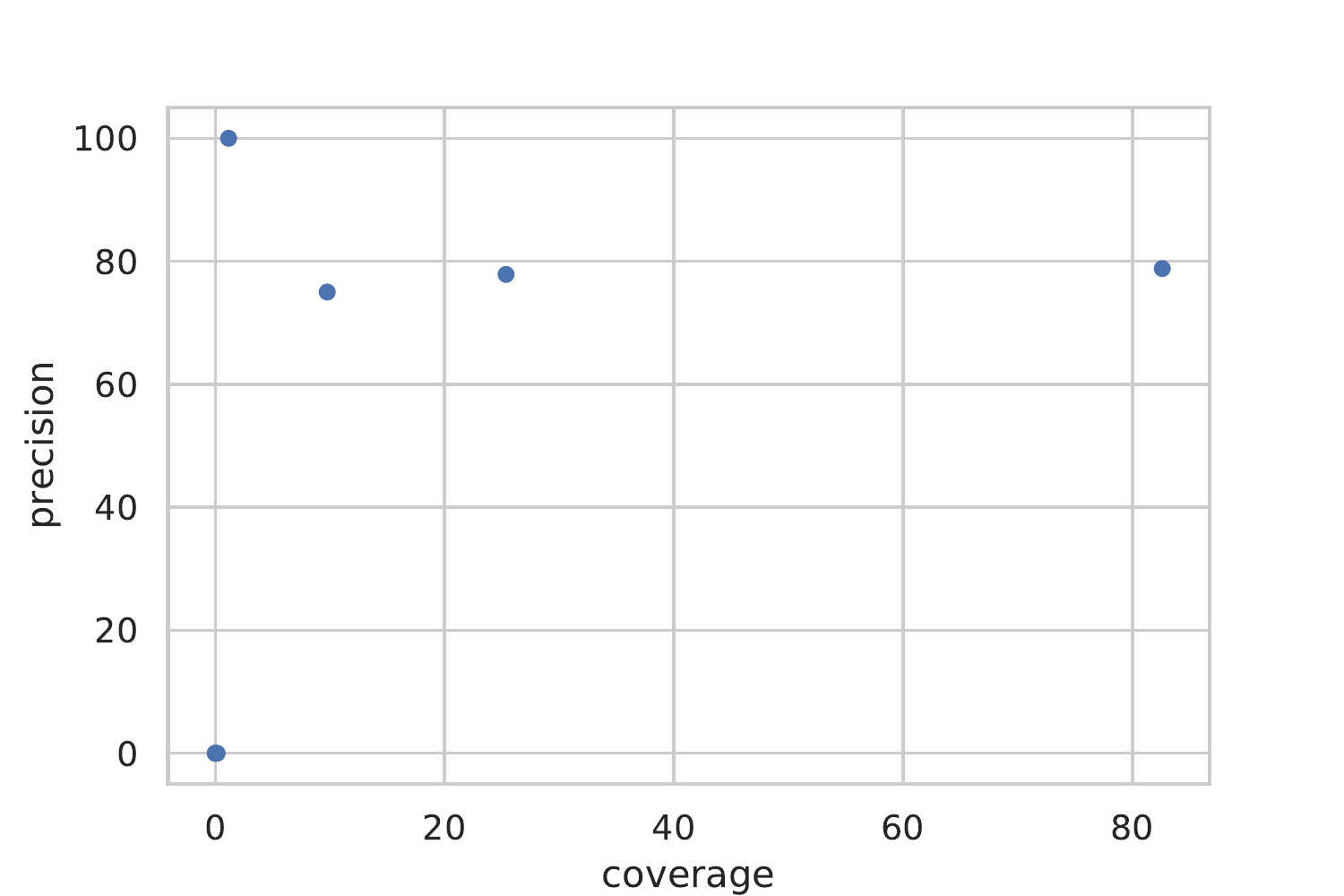}
        \caption{IMDB.}
        \label{fig:rule-precision-recall-plot-imdb}
        \end{subfigure}
        \begin{subfigure}[t]{0.495\linewidth}
        \includegraphics[width=\linewidth]{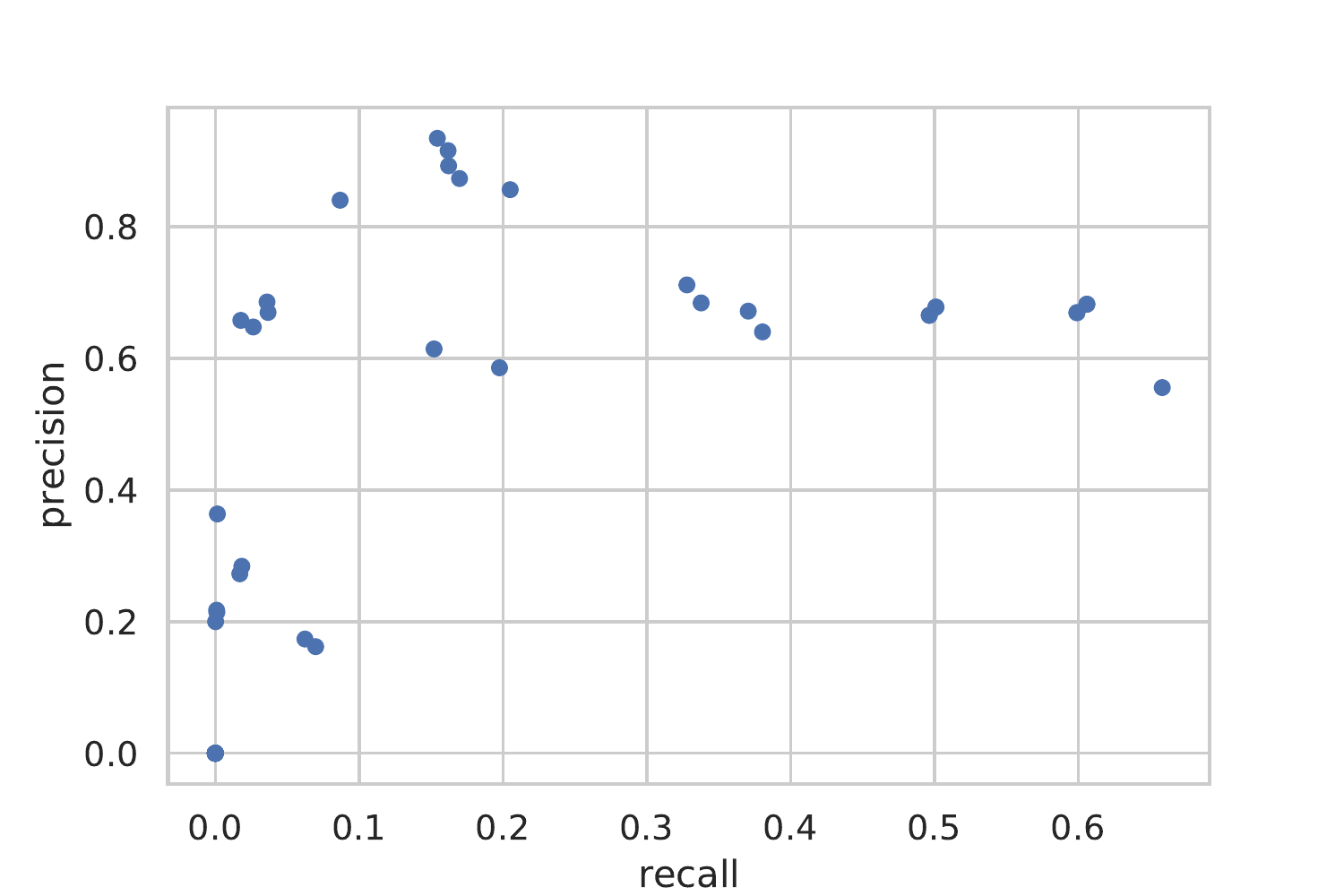}
        \caption{CoNLL.}
        \label{fig:rule-precision-recall-plot-CoNLL}
        \end{subfigure}
        \begin{subfigure}[t]{0.325\linewidth}
        \includegraphics[width=\linewidth]{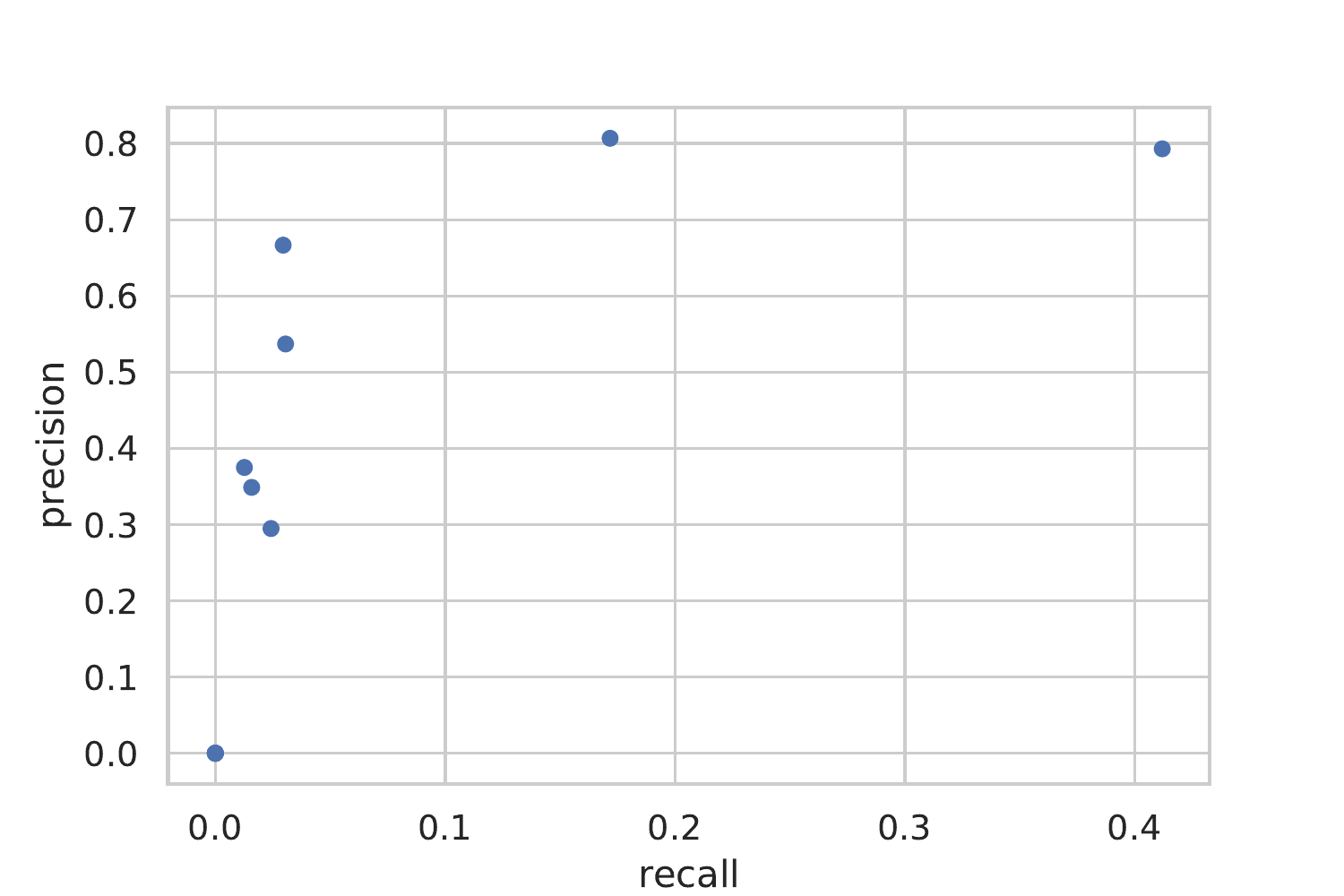}
        \caption{NCBI.}
        \label{fig:rule-precision-recall-plot-NCBI}
        \end{subfigure}
        \begin{subfigure}[t]{0.325\linewidth}
        \includegraphics[width=\linewidth, height=0.67\linewidth]{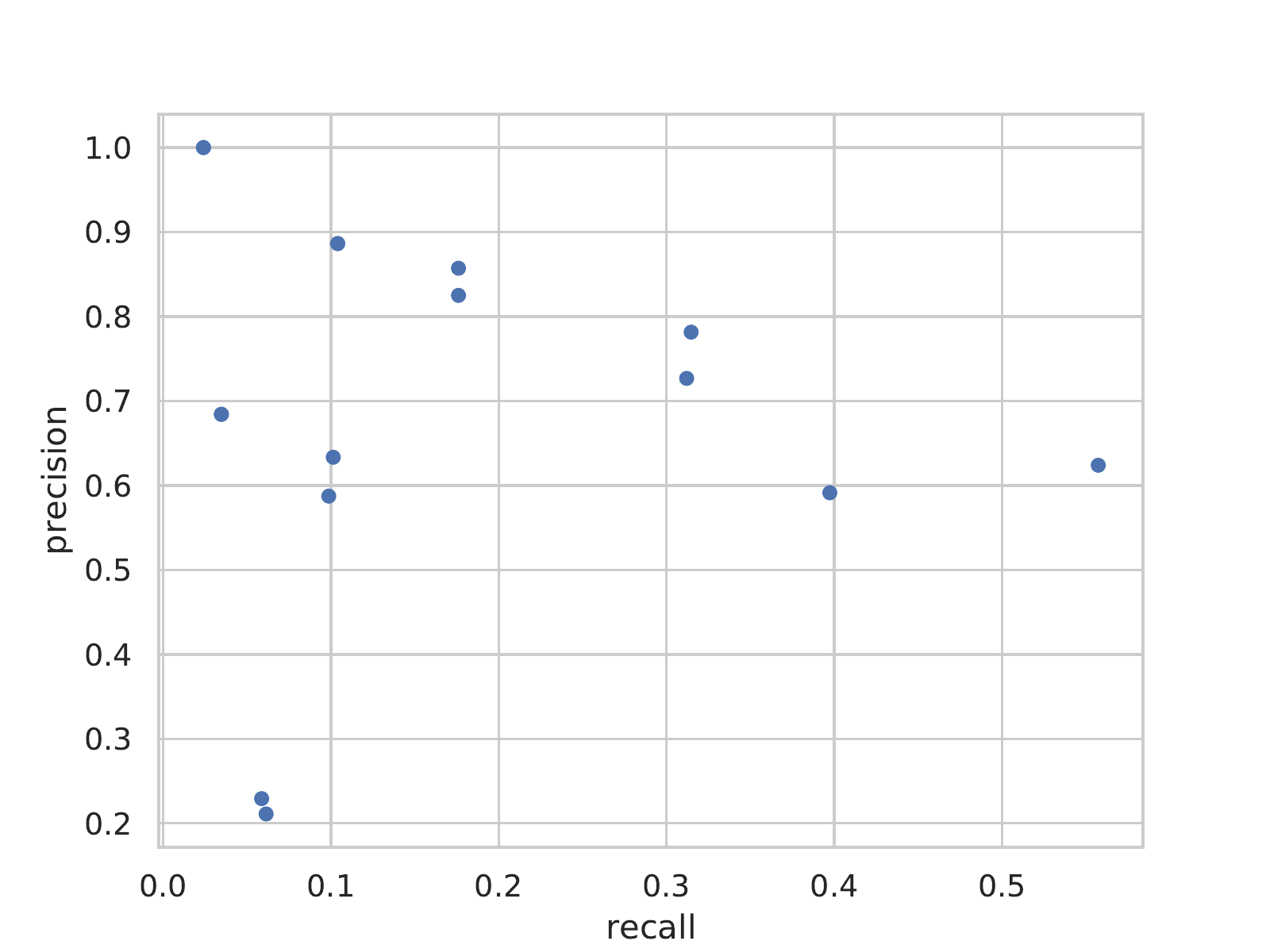}
        \caption{WikiGold.}
        \label{fig:rule-precision-recall-plot-BC5CDR}
        \end{subfigure}
        \begin{subfigure}[t]{0.325\linewidth}
        \includegraphics[width=\linewidth]{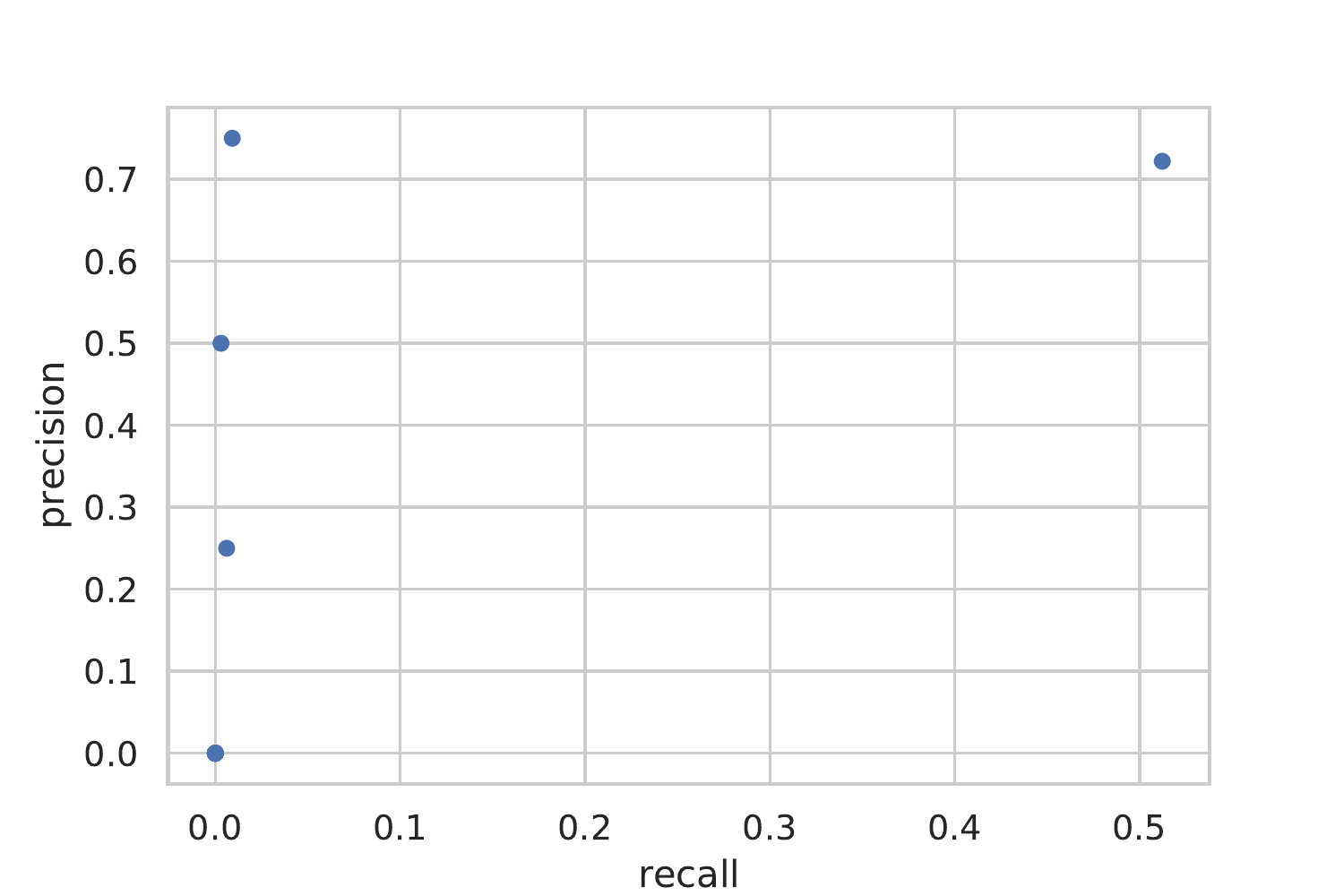}
        \caption{LaptopReview.}
        \label{fig:rule-precision-recall-plot-LaptopReview}
        \end{subfigure}
    \caption{Scatterplots of precision-recall for weak supervision rules. Each point corresponds to a rule. (GossipCop is omitted as it contains only three rules.)}
    \label{fig:rule-precision-recall-plots}
\end{figure*}
\paragraph{Analysis of weak rules.}
For now, we have focused on the evaluation of base models trained using weak labels generated by multiple weak labeling rules. 
It is interesting also to decouple the base model performance from the rule aggregation technique (e.g., majority voting, Snorkel) that was used to generate the training labels, which is an essential modeling component for weak supervision.  
The bottom row in Table~\ref{tab:walnut-main-results} (``Rules'') reports the test performance of rules computed by taking the majority voting of weak labels on the test instances.
(For test instances that are not covered by any rules,  a random class is predicted.) 
 Such majority label is available in our pre-processed datasets.
Interestingly, ``Rules'' sometimes outperforms base models trained using weak labels (``W'', ``Snorkel'').
Note however that ``Rules'' assumes \emph{access to all weak labels} on the test set, which might not always be available. 
On the other hand, the base model learns text features beyond the heuristic-based rules and does not require access to rules during test time and thus can be applied for any test instance.

For a more in-depth analysis of the rule quality,  \name also supports the analysis of individual rules and multi-source aggregation techniques, such as majority voting or Snorkel.
Figure~\ref{fig:rule-precision-recall-plots} shows a precision-recall scatter plot for each rule on each of the dataset in \name. For instance, in the CoNLL dataset rules vary in characteristics, where most rules have a relatively low recall while there are a few rules that have substantially higher recall than the rest.
Across datasets, we observe that rules have higher precision than recall, as most rules are sparse, i.e., apply to a small number of instances in the dataset (e.g., instances containing a specific keyword). Similar trends are observed on other datasets as well.
For detailed descriptions of all weak rules in all datasets, refer to Table~\ref{tab:AGNews_rule_info} -~\ref{tab:LaptopReview_rule_info} in the appendix.

\section{Conclusions}
Motivated by the lack of a unified evaluation platform for semi-weakly
supervised learning for low-resource NLU, in this paper we propose a
new benchmark \name covering a broad range of data domains to advocate
research on leveraging both weak supervision and few-shot clean supervision. We evaluate a series of
different semi-weakly supervised learning methods with different model architecture
on both document-level and token-level classification tasks, and
demonstrate the utility of weak supervision in real-world NLU
tasks. We find that no single semi-weakly supervised learning method wins on \name and there is still gap between semi-weakly supervised learning and fully supervised training.
We expect \name to enable systematic evaluations of
semi-weakly supervised learning methods and stimulate further research in directions
such as more effective learning paradigms leveraging
weak supervision.

\clearpage
\section*{Acknowledgements}

We would like to thank Yichuan Li for helping adapt several weak supervision baselines for document-level classification tasks, and the reviewers for constructive feedback.

\bibliography{anthology,custom, ref}
\newpage 
\clearpage
\appendix

\section{Appendix}
\label{sec:appendix}

\subsection*{Limitations and Broader Impact}

The proposed benchmark is likely to stimulate research on weakly
supervised learning for NLU, and offer the research community on
weakly supervised learning a unified testbed for evaluating new
methodologies developed for low-resource NLU. For many practical NLU
applications, large amount of manually labeled data is unavailable or
expensive to obtain due to either cost or privacy concerns, resorting
to proxy signals such as weak supervision is a viable solution to
mitigate this annotation scarce problem. We hope \name would provide
such an evaluation environment to advocate progress in this direction.

\paragraph{Limitations.}
Due to the lack of existing real-world weak supervision for many NLU
tasks, \name does not include NLU tasks such as Natural Language
Inference for which it is hard to construct weak supervision
rules. Also, currently \name only considers English data; a possible
extension is to also include multi-lingual corpus with weak
supervision available to boost the performance of multi-lingual
language models with weakly supervised learning.

\subsection{Implementation Details}
\label{s:imple_detail}
We implement all baseline experiments with PyTorch and each experiment runs on a single NVIDIA GPU. %
Below are hyper-parameter specifications for all baseline methods (hyperparameters not mentioned below are given default values):
\begin{itemize}
\item Full Clean, C, C+W, Snorkel, C+Snorkel: Batch size is 32 for document-level classification datasets and 16 for the token-level classification datasets. The code for Snorkel is adapted from: \url{https://github.com/snorkel-team/snorkel}. Each training experiment is conducted for the 10 epochs with the checkpoint with the best validation performance saved for evaluation on the test set. 

\item GLC: Code is adapted from \url{https://github.com/mmazeika/glc}. Batch size is 16 for the 4 document-level classification datasets and 8 for the 4 token-level classification datasets. Each experiment trains for 10 epochs with the checkpoint with the best validation performance saved for evaluation on test set.
\item MetaWN: Code is adapted from \url{https://github.com/xjtushujun/meta-weight-net}. Batch size is 8 for the 4 document-level classification datasets and 4 for the 4 token-level classification datasets. The meta-network is a three-layer feed-forward network with hidden dimension of 128. Each experiment trains for 10 epochs with the checkpoint with the best validation performance saved for evaluation on test set.
\item MLC: Code is adapted from \url{https://github.com/microsoft/MLC}. Batch size is 8 for the 4 document-level classification datasets and 4 for the 4 token-level classification datasets. The meta-network is a three-layer feed-forward network with hidden dimension of 128; the label embedding dimension used in the meta-network is 64. Each experiment trains for 10 epochs with the checkpoint with the best validation performance saved for evaluation on test set.
\end{itemize}

To encode input text, we experiment with various text encoders, ranging from
shallow LSTMs to large pre-trained transformerbased encoders~\citep{vaswani2017attention}:
\begin{itemize}
\item BiLSTM-based encoder: the BiLSTM implementations
are all based on 50-dimensional pre-trained glove word embeddings~\citep{pennington2014glove} and bi-directional LSTMs with hidden size 128. Note that our implementation is different than other BiLSTM implementations used by previous work, which are based on 100-dimensional word embeddings and LSTM hidden size 300. This renders our BiLSTM models roughly 3 times smaller than those used by previous work, thus the numbers are not directly comparable. We chose a smaller model capacity for
BiLSTMs to contrast the performance with larger models including DistilBERT and others to show the importance of model capacity on WALNUT. During training, we use a learning rate of 0.005 for all BiLSTM-based models.
\item Transformer-based encoders: we consider pre-trained DistilBERT~\cite{sanh2019distilbert}, BERT~\cite{devlin2018bert}, RoBERTa~\cite{liu2019roberta}, BERT-large, and RoBERTa-large. We fine-tune these models (via the huggingface library) using task-specific classification heads on top of the encoder and a learning rate of 0.00001.
\end{itemize}

\section{Additional Benchmark Details}

\subsection{Document-level classification}
\begin{itemize}
    \item \textbf{AGNews:} and multi-class topic classification (world vs. sports vs. business vs. sci/tech) on news articles from the AGNews dataset~\citep{zhang2015character}. 
    \item \textbf{Yelp:} binary sentiment classification (negative vs. positive) of Yelp restaurant reviews~\citep{zhang2015character}.
    \item \textbf{IMDB:} binary sentiment classification (negative vs. positive) of IMDB movie reviews~\citep{maas2011learning}.
    \item \textbf{GossipCop:} binary fake news detection (fake vs. not fake) on news articles from the GossipCop\footnote{\url{https://www.gossipcop.com/}} fact-checking websites. The GossipCop dataset is part of the fake news detection benchmark FakeNewsNet~\cite{shu2020fakenewsnet}. (We only include the results of Gossipcop to represent fake news classification task as the results for Politifact are similar.)
\end{itemize}

\subsection{Token-level classification}
According to the BIO tagging scheme, “B,” “I,” and “O,” represent the beginning, inside, and outside, of a named entity span, respectively. (Not extracting any values corresponds to a sequence of “O”-only tags.)
Consider, for example, named entity recognition in the CoNLL dataset: 
\begin{table}[!htb]
\small
    \centering
    \begin{tabular}{llllll}
    \textbf{Tokens}:     &  Barack & Obama & lives & in & Washington \\
    \textbf{Tags}:     & B-PER & I-PER & O & O& B-LOC
    \end{tabular}
\end{table}

\begin{itemize}
    \item \textbf{CoNLL:} the CoNLL 2003 dataset~\citep{sang2003introduction} contains news articles from Reuters (split into sentences). In total, there are 35,089 entities from 4 types: organization (ORG), person (PER), location (LOC), and miscellaneous (MISC). Tag classes $C'$: ['O', 'B-PER', 'I-PER', 'B-ORG', 'I-ORG', 'B-LOC', 'I-LOC', 'B-MISC', 'I-MISC']
    \item \textbf{NCBI:} the NCBI Disease corpus~\citep{dougan2014ncbi} contains PubMed abstracts with 6,866 disease mentions. Tag types: ['O', 'B', 'I']
    \item \textbf{WikiGold:} the WikiGold dataset~\citep{balasuriya2009named} contains English Wikipedia articles that were randomly selected and manually annotated with the same entity types as CoNLL. Tag classes $C'$: ['O', 'B-PER', 'I-PER', 'B-ORG', 'I-ORG', 'B-LOC', 'I-LOC', 'B-MISC', 'I-MISC']
    \item \textbf{LaptopReview:} the Laptop Review corpus from the SemEval 2014 Challenge~\citep{pontiki2016semeval} contains 3,012 mentions to laptop features. Tag types $C'$: ['O', 'B', 'I']
\end{itemize}

\begin{figure*}[!htb]
    \centering
    \includegraphics[scale=0.5]{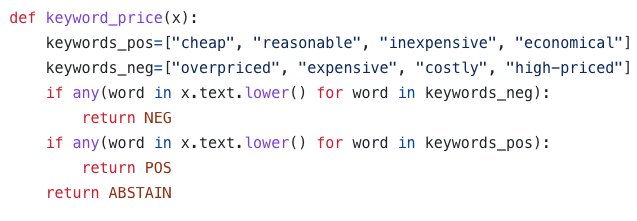}
    \caption{Example of weak rule from the Yelp dataset (rule 6: keyword\_price from Table~\ref{tab:Yelp_rule_info}).}
    \label{fig:ruleEx-Yelp-price}
\end{figure*}

\begin{figure*}[!htb]
    \centering
    \includegraphics[scale=0.5]{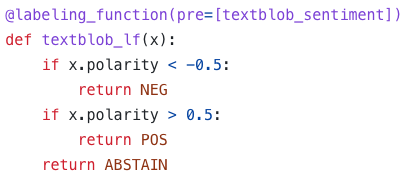}
    \caption{Example of weak rule from the Yelp dataset (rule 1: textblob\_lf from Table~\ref{tab:Yelp_rule_info}).}
    \label{fig:ruleEx-Yelp-textblob}
\end{figure*}

\begin{figure*}[!htb]
    \centering
    \includegraphics[scale=0.4]{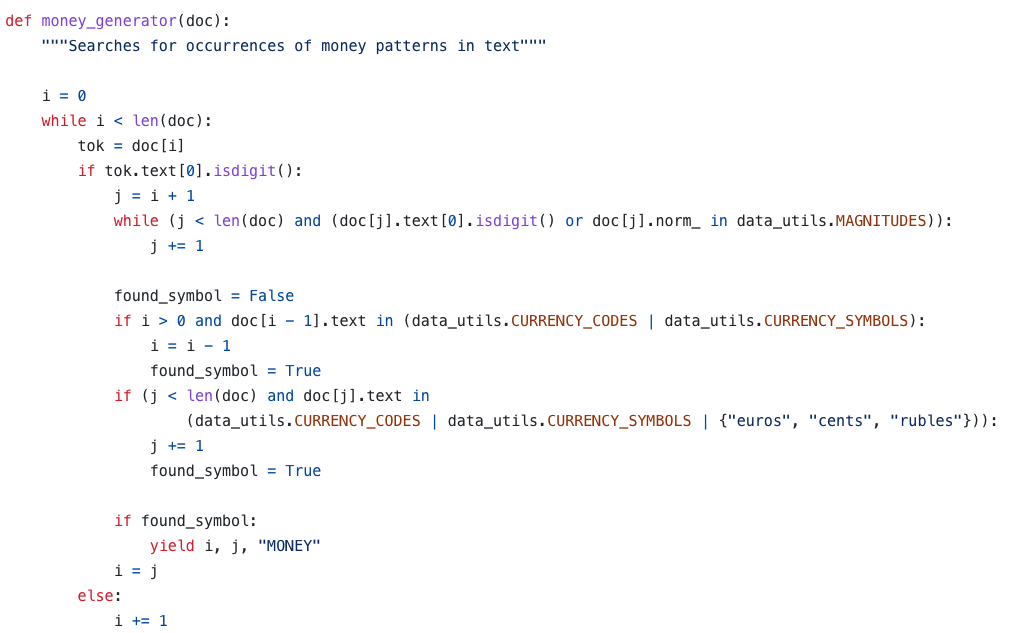}
    \caption{Example of weak rule from the CoNLL dataset (rule 3: money\_detector from Table~\ref{tab:conll_rule_info}). This rule heuristically detects entities that are relevant to money.}
    \label{fig:ruleEx-CoNLL-moneydetector}
\end{figure*}

\begin{figure*}[!htb]
    \centering
    \includegraphics[scale=0.4]{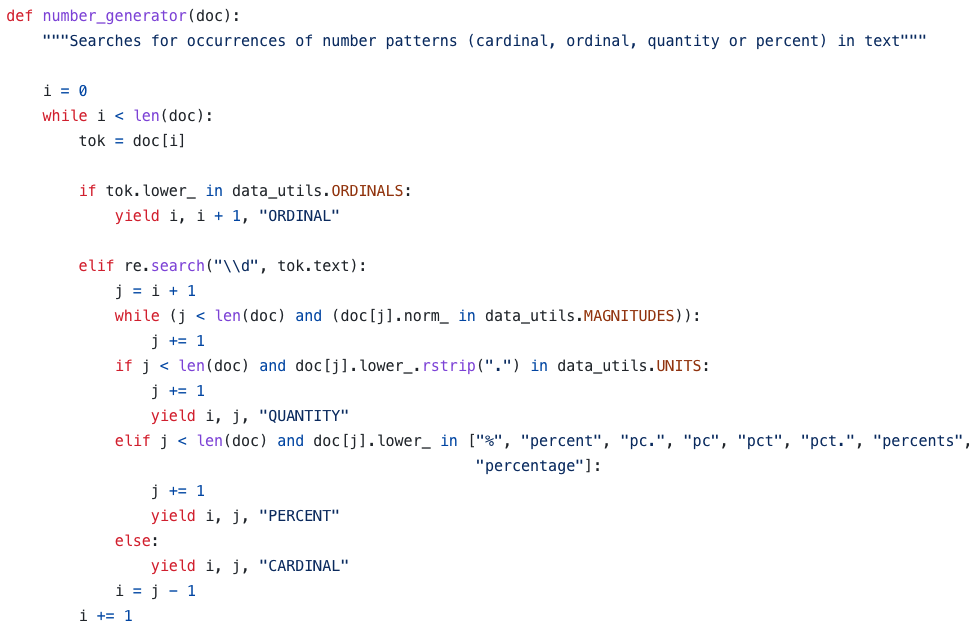}
    \caption{Example of weak rule from the CoNLL dataset (rule 16: number\_detector from Table~\ref{tab:conll_rule_info}). This rule heuristically detects entities that are relevant to numbers.}
    \label{fig:ruleEx-CoNLL-numberdetector}
\end{figure*}

\begin{figure*}[!htb]
    \centering
    \includegraphics[scale=0.5]{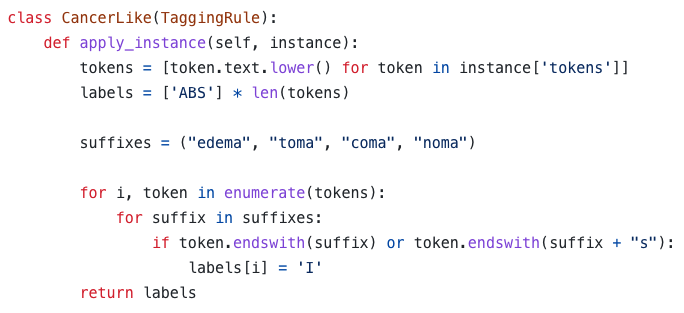}
    \caption{Example of weak rule from the NCBI dataset (rule 3: CancerLike from Table~\ref{tab:NCBI_rule_info}). This rule heuristically detects entities that are relevant to cancer.}
    \label{fig:ruleEx-NCBI-cancerlike}
\end{figure*}

\begin{figure*}[!htb]
    \centering
    \includegraphics[scale=0.5]{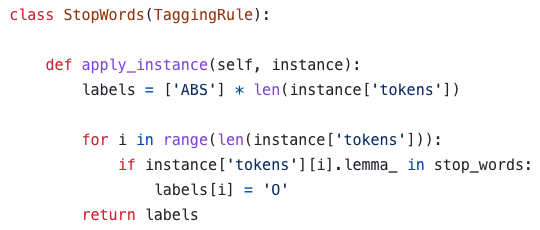}
    \caption{Example of weak rule from the NCBI dataset (rule 11: StopWords from Table~\ref{tab:NCBI_rule_info}). This rule heuristically detects stop words and assigns the `O' tag to the corresponding tokens by assuming that they are not relevant to any disease.}
    \label{fig:ruleEx-NCBI-stopwords}
\end{figure*}

\begin{figure*}[!htb]
    \centering
    \includegraphics[scale=0.5]{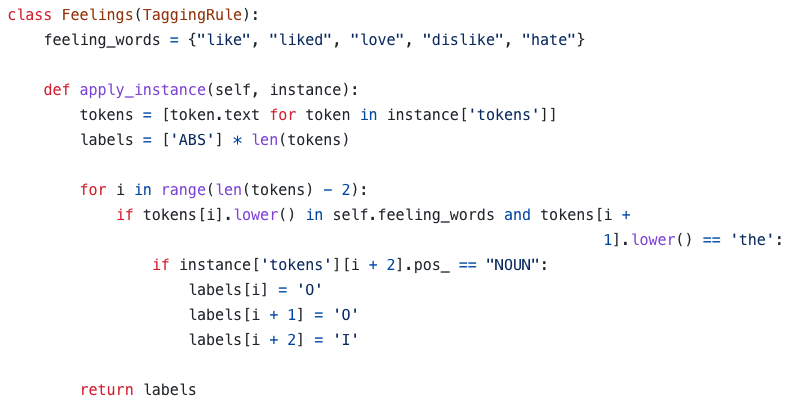}
    \caption{Example of weak rule from the LaptopReview dataset (rule 5: Feelings from Table~\ref{tab:LaptopReview_rule_info}). This rule heuristically detects entities that are relevant to laptop features based on keywords that express the user's feelings.}
    \label{fig:ruleEx-Laptopreview-feelings}
\end{figure*}

Table~\ref{tab:token-level-dataset-stats-detailed} shows detailed statistics for token-level classification datasets.
More dataset statistics are provided in Table~\ref{tab:doc_data}.
Tables~\ref{tab:AGNews_rule_info}-\ref{tab:LaptopReview_rule_info} show detailed information for all rules.
Figures~\ref{fig:ruleEx-Yelp-price}-\ref{fig:ruleEx-Laptopreview-feelings} show examples of weak rules for various datasets.

\begin{table*}[!htb]
    \centering
    \caption{Extra token-level statistics for the token-level classification datasets.}
    \begin{tabular}{l|l|l|l|l}
    \toprule
         & \textbf{CoNLL} & \textbf{NCBI} & \textbf{WikiGold} & \textbf{LaptopReview} \\
    \midrule 
    \# train tokens & 203,621 &  135,572 & 31,560 & 41,525\\
    \# dev tokens & 51,362 & 23,789 &  3,683& 9,970\\
    \# test tokens & 46,435 & 24,219 &  3,762& 11,884\\
    \bottomrule
    \end{tabular}
    \label{tab:token-level-dataset-stats-detailed}
\end{table*}

\begin{table*}[]
    \centering
    \caption{List of rules for the AGNews dataset. The rules are the same as the tagging rules in~\cite{zhang2015character}. The Python implementations can be found in: {\small \url{https://github.com/weakrules/Denoise-multi-weak-sources/blob/master/rules-noisy-labels/Agnews/angews_rule.py}}}
    \resizebox{0.6\linewidth}{!}{
\begin{tabular}{ll} \\
 \textbf{Rule name} & \textbf{Description} \\  
 \toprule 
1. \labelfunsmall{world1} & Keyword-based detection of the \nerlabelto{WORLD} topic\\
2.   \labelfunsmall{world2} & Keyword-based detection of the \nerlabelto{WORLD} topic\\
3. \labelfunsmall{sports1} & Keyword-based detection of the \nerlabelto{SPORTS} topic\\
4. \labelfunsmall{sports2} & Keyword-based detection of the \nerlabelto{SPORTS} topic\\
5. \labelfunsmall{sports3} & Keyword-based detection of the \nerlabelto{SPORTS} topic\\
6. \labelfunsmall{tech1} & Keyword-based detection of the \nerlabelto{TECH} topic\\
7. \labelfunsmall{tech2} & Keyword-based detection of the \nerlabelto{TECH} topic\\
8. \labelfunsmall{business1} & Keyword-based detection of the \nerlabelto{BUSINESS} topic\\
9. \labelfunsmall{business2} & Keyword-based detection of the \nerlabelto{BUSINESS} topic\\
\bottomrule
\end{tabular}}
\label{tab:AGNews_rule_info}
\end{table*}

\begin{table*}[]
    \centering
    \caption{List of rules for the IMDB dataset. The rules are the same as in~\cite{zhang2015character}. The Python implementations can be found in: {\small\url{https://github.com/weakrules/Denoise-multi-weak-sources/blob/master/rules-noisy-labels/IMDB/imdb_rule.py}}}
    \resizebox{\linewidth}{!}{
\begin{tabular}{ll} \\
 \textbf{Rule name} & \textbf{Description} \\  
 \toprule 
1. \labelfunsmall{expression\_nexttime} & Regex-based detection of \nerlabelto{POSITIVE} sentiment (re-watching expressions)\\
2. \labelfunsmall{expression\_recommend} & Regex-based detection of \nerlabelto{POSITIVE} sentiment (recommendation expressions)\\
3. \labelfunsmall{expression\_value} & Regex-based detection of \nerlabelto{POSITIVE} sentiment (value expressions)\\
4.   \labelfunsmall{keyword\_compare} & Keyword-based detection of \nerlabelto{NEGATIVE} sentiment based on movie comparisons\\
5. \labelfunsmall{keyword\_general} & Keyword-based detection of \nerlabelto{POSITIVE} and \nerlabelto{NEGATIVE} sentiment\\
6.   \labelfunsmall{keyword\_actor} & Keyword-based detection of \nerlabelto{POSITIVE} sentiment regarding the actors\\
7.   \labelfunsmall{keyword\_finish} & Keyword-based detection of \nerlabelto{NEGATIVE} sentiment\\
8.   \labelfunsmall{keyword\_plot} & Keyword-based detection of \nerlabelto{POSITIVE} and \nerlabelto{NEGATIVE} sentiment regarding the plot\\
\bottomrule
\end{tabular}}

\label{tab:IMDB_rule_info}
\end{table*}

\begin{table*}[]
    \centering
\caption{List of rules for the Yelp dataset. The rules are the same as in~\cite{zhang2015character}. The Python implementations can be found in: {\small\url{https://github.com/weakrules/Denoise-multi-weak-sources/blob/master/rules-noisy-labels/Yelp/yelp_rules.py}}}    
    \resizebox{\linewidth}{!}{
\begin{tabular}{ll} \\
 \textbf{Rule name} & \textbf{Description} \\  
 \toprule 
1. \labelfunsmall{textblob\_lf} & Model-based detection of \nerlabelto{POSITIVE} and \nerlabelto{NEGATIVE} sentiment (TextBlob model)\\
2. \labelfunsmall{keyword\_recommend} & Regex-based detection of \nerlabelto{POSITIVE}  sentiment (recommendation expressions)\\
3. \labelfunsmall{keyword\_general} & Regex-based detection of \nerlabelto{POSITIVE} and \nerlabelto{POSITIVE} sentiment (general expressions)\\
4.   \labelfunsmall{keyword\_mood} & Keyword-based detection of \nerlabelto{POSITIVE} and \nerlabelto{NEGATIVE} sentiment based on the user's mood\\
5. \labelfunsmall{keyword\_service} & Keyword-based detection of \nerlabelto{POSITIVE} and \nerlabelto{NEGATIVE} sentiment relevant to the service\\
6.   \labelfunsmall{keyword\_price} & Keyword-based detection of \nerlabelto{POSITIVE} and \nerlabelto{NEGATIVE} sentiment regarding the prices\\
7.   \labelfunsmall{keyword\_environment} & Keyword-based detection of \nerlabelto{POSITIVE} and \nerlabelto{NEGATIVE} sentiment relevant to the ambience\\
8.   \labelfunsmall{keyword\_food} & Keyword-based detection of \nerlabelto{POSITIVE} and \nerlabelto{NEGATIVE} sentiment relevant to the food\\

\bottomrule
\end{tabular}}

\label{tab:Yelp_rule_info}
\end{table*}

\begin{table*}[]
    \centering
\caption{List of rules for the GossipCop dataset. The rules are the same as in~\cite{shu2020fakenewsnet} (page 8 in {\small \url{http://www.cs.iit.edu/~kshu/files/ecml_pkdd_mwss.pdf}).}}    
    \resizebox{\linewidth}{!}{
\begin{tabular}{ll} \\
 \textbf{Rule name} & \textbf{Description} \\  
 \toprule 
1. \labelfunsmall{mean\_scores} & 
User interaction-based detection of \nerlabelto{FAKE} news: If a news piece has a\\
&  standard deviation of user sentiment scores greater than a threshold $\tau_1$,\\
& then the news is weakly labeled as \nerlabelto{FAKE} news. \\
2. \labelfunsmall{std\_scores} & 
User interaction-based detection of \nerlabelto{FAKE} news: If the mean value of \\
& users' absolute bias scores - sharing a piece of news – is greater than \\
& a threshold $\tau_2$, then the news piece is weakly-labeled as \nerlabelto{FAKE} news. \\
3. \labelfunsmall{credibility\_score} & 
User interaction-based detection of \nerlabelto{FAKE} news: If a news piece has \\
& an average credibility score less than a threshold $\tau_3$, then the news \\
& is weakly-labeled as \nerlabelto{FAKE} news. \\

\bottomrule
\end{tabular}}

\label{tab:GossipCop_rule_info}
\end{table*}

\begin{table*}[]
    \centering
\caption{List of rules for the CoNLL dataset. The Python implementation of CoNLL rules is provided in the ``skweak'' repo: {\small \url{https://github.com/NorskRegnesentral/skweak/blob/670fcdec680930ce3e497886d06d61e6a1f2c195/examples/ner/conll2003_ner.py}}}
    \resizebox{\linewidth}{!}{
\begin{tabular}{ll} \\
 \textbf{Rule name} & \textbf{Description} \\  
 \toprule 
1. \labelfunsmall{date\_detector} & Heuristic detection of entities of type \nerlabelto{DATE} \\
2. \labelfunsmall{time\_detector}  &  Heuristic detection of entities of type \nerlabelto{TIME}  \\
3. \labelfunsmall{money\_detector} & Heuristic detection of entities of type \nerlabelto{MONEY} \\
4. \labelfunsmall{proper\_detector}  & Heuristic detection of proper names based on casing  \\
 5. \labelfunsmall{infrequent\_proper\_detector} &  Heuristic detection of proper names based on casing \\  & +  including at least one infrequent token \\
 6. \labelfunsmall{proper2\_detector}  & Heuristic detection of proper names based on casing  \\
 7. \labelfunsmall{infrequent\_proper2\_detector} & Heuristic detection of proper names based on casing \\  & +  including at least one infrequent token \\
 8. \labelfunsmall{nnp\_detector} & Heuristic detection of sequences of tokens with \nerlabelto{NNP} as POS-tag\\
 9. \labelfunsmall{infrequent\_nnp\_detector} & Heuristic detection of sequences of tokens with \nerlabelto{NNP} as POS-tag \\  & + including at least one infrequent token (rank $>$ 15000 in vocabulary)   \\
 10. \labelfunsmall{compound\_detector} & Heuristic detection of proper noun phrases with compound dependency relations \\
 11. \labelfunsmall{infrequent\_compound\_detector} & Heuristic detection of proper noun phrases with compound dependency relations \\  & +  including at least one infrequent token  \\
 12. \labelfunsmall{misc\_detector} &  Heuristic detection of entities of type \nerlabelto{NORP}, \nerlabelto{LANGUAGE}, \nerlabelto{FAC} or \nerlabelto{EVENT} \\
13. \labelfunsmall{legal\_detector} & Heuristic detection of entities of type \nerlabelto{LAW} \\
14. \labelfunsmall{company\_type\_detector}& Gazetteer using a large list of company names  \\

 15. \labelfunsmall{full\_name\_detector} & Heuristic function to detect full person names \\
 16. \labelfunsmall{number\_detector} & Heuristic detection of entities \nerlabelto{CARDINAL},\nerlabelto{ORDINAL}, \nerlabelto{PERCENT} and \nerlabelto{QUANTITY} \\
 17. \labelfunsmall{snips}  & Probabilistic parser specialised in the recognition of dates,\\ & + times, money amounts, percents, and cardinal/ordinal values \\
18. \labelfunsmall{core\_web\_md} &  NER model trained on Ontonotes 5.0  \\
19. \labelfunsmall{core\_web\_md+c} & NER model trained on Ontonotes 5.0 + postprocessing \\
 20. \labelfunsmall{BTC} &  NER model trained on the Broad Twitter Corpus \\
 21. \labelfunsmall{BTC+c}& NER model trained on the Broad Twitter Corpus + postprocessing  \\
 22. \labelfunsmall{SEC}&  NER model trained on SEC-filings \\
 23. \labelfunsmall{SEC+c} & NER model trained on SEC-filings + postprocessing \\
24. \labelfunsmall{edited\_core\_web\_md} & NER model trained on Ontonotes 5.0 + alternative postprocessing\\
25. \labelfunsmall{edited\_core\_web\_md+c} & NER model trained on Ontonotes 5.0 + alternative postprocessing\\
 26. \labelfunsmall{wiki\_cased}  & Gazetteer (case-sensitive) using Wikipedia entries \\
 27. \labelfunsmall{wiki\_uncased} & Gazetteer (case-insensitive) using Wikipedia entries \\
 28. \labelfunsmall{multitoken\_wiki\_cased} & Same as above, but restricted to multitoken entities \\

 29. \labelfunsmall{multitoken\_wiki\_uncased} &  Same as above, but restricted to multitoken entities\\ 
 30. \labelfunsmall{wiki\_small\_cased} &  Gazetteer (case-sensitive) using Wikipedia entries with non-empty description\\
 31. \labelfunsmall{wiki\_small\_uncased} & Gazetteer (case-insensitive) using Wikipedia entries with non-empty description \\
 32. \labelfunsmall{multitoken\_wiki\_small\_cased} & Same as above, but restricted to multitoken entities \\

 33. \labelfunsmall{multitoken\_wiki\_small\_uncased} & Same as above, but restricted to multitoken entities \\
 34. \labelfunsmall{geo\_cased} & Gazetteer (case-sensitive) using the Geonames database \\
 35. \labelfunsmall{geo\_uncased}  &  Gazetteer (case-insensitive) using the Geonames database \\
 36. \labelfunsmall{multitoken\_geo\_cased} & Same as above, but restricted to multitoken entities \\
 37. \labelfunsmall{multitoken\_geo\_uncased} & Same as above, but restricted to multitoken entities \\
 38. \labelfunsmall{crunchbase\_cased} & Gazetteer (case-sensitive) using the Crunchbase Open Data Map \\
 39. \labelfunsmall{crunchbase\_uncased} & Gazetteer (case-insensitive) using the Crunchbase Open Data Map \\
 40. \labelfunsmall{multitoken\_crunchbase\_cased} & Same as above, but restricted to multitoken entities \\
 41. \labelfunsmall{multitoken\_crunchbase\_uncased} & Same as above, but restricted to multitoken entities \\
 42. \labelfunsmall{product\_cased}  & Gazetteer (case-sensitive) using products extracted from DBPedia \\
 43. \labelfunsmall{product\_uncased} &  Gazetteer (case-insensitive) using products extracted from DBPedia \\
 44. \labelfunsmall{multitoken\_product\_cased} & Same as above, but restricted to multitoken entities \\
 45. \labelfunsmall{multitoken\_product\_uncased} & Same as above, but restricted to multitoken entities  \\
 46. \labelfunsmall{doclevel\_voter}  & Considers all appearances of the same entity string in the document\\

 47. \labelfunsmall{doc\_history\_cased}  & Considers already introduced entities in the document (case-sensitive)\\
 48. \labelfunsmall{doc\_history\_uncased}  & Considers already introduced entities in the document (case-insensitive)\\
 49. \labelfunsmall{doc\_majority\_cased} &  Considers all entities in the document (case-sensitive) \\
 50. \labelfunsmall{doc\_majority\_uncased} &  Considers all majority labels in the document (case-insensitive) \\
\bottomrule

\end{tabular}}

    \label{tab:conll_rule_info}
\end{table*}

\begin{table*}[!htb]
    \centering
\caption{List of rules for the NCBI dataset. The rules are the same as the tagging rules in~\cite{safranchik2020weakly}. Python implementations: {\small \url{https://github.com/BatsResearch/safranchik-aaai20-code/blob/master/NCBI-Disease/train_generative_models.py}}}    
    \resizebox{0.9\linewidth}{!}{
\begin{tabular}{ll} \\
 \textbf{Rule name} & \textbf{Description} \\  
 \toprule 
1. \labelfunsmall{CoreDictionaryUncased} & AutoNER dictionary (biomedical entities) \\
2. \labelfunsmall{CoreDictionaryExact} & AutoNER dictionary (biomedical entities, exact match)\\
3. \labelfunsmall{CancerLike} & Heuristic detection of entities that are relevant to cancer\\
4. \labelfunsmall{CommonSuffixes} & Heuristic detection of entities that are relevant to common diseases\\
5. \labelfunsmall{Deficiency} & Heuristic detection of entities that are relevant to deficiencies\\
6. \labelfunsmall{Disorder} & Heuristic detection of entities that are relevant to disorders\\
7. \labelfunsmall{Lesion} & Heuristic detection of entities that are relevant to lesions\\
8. \labelfunsmall{Syndrome} & Heuristic detection of entities that are relevant to syndroms\\
9. \labelfunsmall{BodyTerms} & UMLS dictionary entries for terms that are relevant to body parts\\
10. \labelfunsmall{OtherPOS} & Heuristic detection of parts of speech that are not relevant to any disease\\
11. \labelfunsmall{StopWords} & Heuristic detection of stop words that are not relevant to any disease\\
12. \labelfunsmall{Punctuation} & Heuristic detection of punctiations that are not relevant to any disease\\
\bottomrule

\end{tabular}}

\label{tab:NCBI_rule_info}
\end{table*}

\begin{table*}[!htb]
    \centering
\caption{List of rules for the WikiGold dataset. The Python implementation of WikiGold rules is provided in the ``skweak'' repo: {\small \url{https://github.com/NorskRegnesentral/skweak/blob/670fcdec680930ce3e497886d06d61e6a1f2c195/examples/ner/conll2003_ner.py}}}    
    \resizebox{\linewidth}{!}{
\begin{tabular}{ll} \\
 \textbf{Rule name} & \textbf{Description} \\  
 \toprule 
  1. \labelfunsmall{BTC} &  NER model trained on the Broad Twitter Corpus \\
2. \labelfunsmall{core\_web\_md} &  NER model trained on Ontonotes 5.0  \\
 3. \labelfunsmall{crunchbase\_cased} & Gazetteer (case-sensitive) using the Crunchbase Open Data Map \\
 4. \labelfunsmall{crunchbase\_uncased} & Gazetteer (case-insensitive) using the Crunchbase Open Data Map \\
 5. \labelfunsmall{full\_name\_detector} & Heuristic function to detect full person names \\
 6. \labelfunsmall{geo\_cased} & Gazetteer (case-sensitive) using the Geonames database \\
 7. \labelfunsmall{geo\_uncased}  &  Gazetteer (case-insensitive) using the Geonames database \\
  8. \labelfunsmall{misc\_detector} &  Heuristic detection of entities of type \nerlabelto{NORP}, \nerlabelto{LANGUAGE}, \nerlabelto{FAC} or \nerlabelto{EVENT} \\
  9. \labelfunsmall{wiki\_cased}  & Gazetteer (case-sensitive) using Wikipedia entries \\
 10. \labelfunsmall{wiki\_uncased} & Gazetteer (case-insensitive) using Wikipedia entries \\
  11. \labelfunsmall{multitoken\_crunchbase\_cased} & Same as above, but restricted to multitoken entities \\
 12. \labelfunsmall{multitoken\_crunchbase\_uncased} & Same as above, but restricted to multitoken entities \\
 13. \labelfunsmall{multitoken\_geo\_cased} & Same as above, but restricted to multitoken entities \\
 14. \labelfunsmall{multitoken\_geo\_uncased} & Same as above, but restricted to multitoken entities \\
 15. \labelfunsmall{multitoken\_wiki\_cased} & Same as above, but restricted to multitoken entities \\
 16. \labelfunsmall{multitoken\_wiki\_uncased} &  Same as above, but restricted to multitoken entities\\ 
\end{tabular}}
\label{tab:wikigold_rule_info}
\end{table*}

\begin{table*}[!htb]
    \centering
\caption{List of rules for the LaptopReview dataset. The rules are the same as the tagging rules in~\cite{safranchik2020weakly}. Python implementations: {\small \url{https://github.com/BatsResearch/safranchik-aaai20-code/blob/master/LaptopReview/train_generative_models.py}}}    
    \resizebox{0.9\linewidth}{!}{
\begin{tabular}{ll} \\
 \textbf{Rule name} & \textbf{Description} \\  
 \toprule 
1. \labelfunsmall{CoreDictionary} &  AutoNER dict with entries of terms that are relevant to electronics\\
2. \labelfunsmall{OtherTerms} &  Heuristic detection of laptop entities based on a pre-defined keyword list\\
3. \labelfunsmall{ReplaceThe} &  Heuristic detection of laptop entities based on the ``replace the'' phrase\\
4. \labelfunsmall{iStuff} &  Heuristic detection of laptop entities based on uppercase letters\\
5. \labelfunsmall{Feelings} &  Heuristic detection of laptop entities based on common expressions\\
6. \labelfunsmall{ProblemWithThe} &  Heuristic detection of laptop entities based on common expressions\\
7. \labelfunsmall{External} &  Heuristic detection of laptop entities based on common hardware expression\\
8. \labelfunsmall{StopWords} &  Heuristic detection of stop words that are not relevant to electronics\\
9. \labelfunsmall{Punctuation} &  Heuristic detection of punctuation that are not relevant to electronics\\
10. \labelfunsmall{Pronouns} &  Heuristic detection of pronouns that are not relevant to electronics\\
11. \labelfunsmall{NotFeatures} &Heuristic detection of terms that are not relevant to laptop features \\
12. \labelfunsmall{Adv} &  Heuristic detection of adverbs that are not relevant to electronics\\
\bottomrule
\end{tabular}}

    \label{tab:LaptopReview_rule_info}
\end{table*}

\newpage 
\clearpage

\clearpage

\section{Additional Results}

Table~\ref{tab:std-results} shows standard deviation results for all datasets, methods, and base models. 
The rightmost column responds the average standard deviation (AVG std) across tasks, which we also reported in Table~\ref{tab:average-f1-std-results}.

\begin{table*}[ht]
\small
    \centering
    \caption{Standard deviation results on \name.}
    \resizebox{\linewidth}{!}{

    \begin{tabular}{r|cccccccc|c}
    \toprule
    Method & AGNews & IMDB & Yelp  & GossipCop & CoNLL & NCBI & WikiGold & LaptopReview & AVG \\\midrule
    \multicolumn{10}{c}{BiLSTM} \\
    \rowcolor{mygray}    
 Full Clean& 0.2 & 0.5 & 0.3 & 1.0 &7.2&	1.6&	0.5&	2.5 & 1.7 \\
 C & 1.1 & 2.9 & 4.4 & 2.0 &1.0 &	1.5&	0.9& 	5.0 & 2.4 \\
W & 6.1 & 0.5 & 2.4 & 0.9 & 3.2	&1.5&	0.6&	1.2  & 2.1 \\
C+W & 0.3 & 0.9 & 1.2 & 1.1 &6.7 &	1.1	& 0.6&	2.9 &  1.9\\
Snorkel & 3.9 & 0.8 & 0.6 & 0.4 & 2.3 & 2.1	&0.7&	1.4 & 1.5 \\
C+Snorkel & 0.4 & 1.0 & 1.2 & 1.5 & 2.8 & 1.7&	0.6&	2.1 & 1.4 \\
GLC & 1.3 & 0.3 & 0.9 & 1.7 &  7.2&	1.2 &	0.5 &	0.9 & 1.7\\
MetaWN & 1.8 & 0.3 & 2.0 & 1.3 & 0.0 &	1.3 &	0.4 &	0.5 & 0.9\\
MLC & 2.1 & 0.3 & 3.8 & 1.6 & 0.0&	1.1&	0.3	&0.7 & 1.2\\
\hline 
    \multicolumn{10}{c}{DistilBERT} \\
        \rowcolor{mygray}
Full Clean& 0.2 & 0.3 & 0.3 & 1.5 & 0.4 & 0.4 & 0.7 & 3.5 &  0.9\\
 C & 6.2 &6.8  &7.4  & 6.4 & 3.6 & 2.0   & 1.3 & 1.7 &  4.4\\
W & 3.9 & 1.1 & 1.5 & 1.1 & 0.9 & 1.2 & 0.3 & 2.0   &  1.5\\
C+W & 0.6 & 0.2 & 0.9 & 1.2 & 0.8 & 1.4 & 0.4 & 3.6 &  1.1\\
Snorkel & 3.0 & 1.6 & 0.6 & 0.5 & 1.1 & 1.5 & 0.4 & 2.2 &  1.4\\
C+Snorkel &0.7  &0.5  & 2.0 & 0.8 & 0.9 & 1.9 & 0.4 & 3.6 & 1.4 \\
GLC & 2.8 & 0.5 & 1.5 & 2.0 & 2.1&	1.8	&0.2&	1.5 & 1.5\\
MetaWN & 1.6 & 1.3 & 0.8 & 2.2 & 1.8	&1.4&	0.3	&0.6 & 1.3 \\
MLC & 2.5 & 0.6 & 0.7 & 1.6 &  1.2&	1.8	&0.4&	2.3 & 1.4\\
\hline 
\multicolumn{10}{c}{BERT} \\
    \rowcolor{mygray}
Full Clean& 0.1 & 0.5 & 0.2 & 1.0 & 0.6 & 0.5 & 1.0   & 1.8 & 0.7 \\
C & 0.9 & 8.1 & 5.2 & 1.8 & 1.3 & 0.8 & 1.3 & 2.5 &  2.7\\
W &  2.7& 0.5 & 1.1 & 2.2 & 1.2 & 2.8 & 0.9 & 1.8 &  1.7\\
C+W & 0.4 & 0.6 & 1.4 &  1.5& 0.9 & 1.4 & 0.8 & 1.5 &  1.1\\
 Snorkel  & 2.3 &  3.7&1.3  & 0.9 & 1.3 & 3.5 & 1.0   & 1.3 & 1.9 \\
C+Snorkel & 1.0 & 0.5 & 0.6 & 0.6 & 1.6 & 1.7 & 0.8 & 2.6 &  1.2\\
GLC & 1.6 & 0.8 & 2.2 & 2.8 & 2.4&	1.2	&0.4&	1.2 & 1.6 \\
MetaWN & 1.1 & 1.0 & 1.0 & 2.4 & 1.6&	0.5	&0.3&	1.4 & 1.2\\
MLC  &2.0  & 0.8 & 1.3 & 1.4 & 2.1&	2.8	&0.2&	0.4 & 1.4\\
\hline 
\multicolumn{10}{c}{RoBERTa} \\
    \rowcolor{mygray}
Full Clean& 0.1 & 0.4 & 0.2 & 1.0 & 0.3 & 0.7 & 1.0   & 2.0   &  0.7\\
 C& 2.0 & 5.4 & 5.9 & 5.2 & 2.3 & 2.1 & 1.7 & 4.1 &  3.6\\
W & 1.2 & 0.7 & 1.2 & 2.4 & 1.4 & 1.5 & 0.9 & 2.7 &  1.5\\
C+W & 0.9 & 1.7 & 1.4 & 1.0 & 1.6 & 1.5 & 0.6 & 5.3 &  1.8\\
Snorkel  & 3.2 & 2.3 & 2.9 & 0.6 & 2.0 & 1.1 & 0.9 & 2.9 &  2.0\\
C+Snorkel & 0.7 & 2.2 & 1.6 & 1.8 & 1.8 & 3.1 & 0.8 & 5.7 & 2.2\\
GLC & 1.3 & 0.7 & 1.8 & 2.3 & 3.2&	0.4	&0.4&	0.8 & 1.4\\
MetaWN & 2.7 & 0.9 & 1.4 & 2.1 & 0.7&	0.9&	0.3&	1.1 & 1.3\\
MLC & 1.6 & 1.2 & 1.0 & 1.3 & 1.2&	2.7	&0.2&	3.1  & 1.6\\\hline
\end{tabular}}
\resizebox{\linewidth}{!}{
\begin{tabular}{rcccccccc|c}
\vspace{-4mm}
   \textcolor{white}{Method} & \textcolor{white}{AGNews} & \textcolor{white}{IMDB} & \textcolor{white}{Yelp}  & \textcolor{white}{GossipCop} & \textcolor{white}{CoNLL} & \textcolor{white}{NCBI} & \textcolor{white}{BC5CDR} & \textcolor{white}{LaptopReview} & \textcolor{white}{AVG}\\
\multicolumn{10}{c}{BERT-large} \\
    \rowcolor{mygray}
Full Clean& 0.1 & 0.4 & 0.3 & 0.6 &  1.0&1.4&1.2&3.1  & 1.0\\
C&  22.6 & 3.7 & 5.8 & 3.2&  4.0&2.4&2.3&3.4  & 5.9\\
W& 1.1 & 2.4 & 1.2 & 1.4 & 1.2&2.0&1.1&1.0  & 1.4\\
C+W& 2.1 & 1.6 & 0.9 & 1.8 &1.6&1.9&0.9&3.8  &1.8 \\
Snorkel&  2.2& 1.5 & 0.5 & 1.5 & 0.7&4.0&1.1&1.6  & 1.6\\
C+Snorkel& 0.9 & 1.4 & 0.8 & 1.4 & 1.0&4.5&1.1&2.8 &1.7 \\
GLC& 2.0 & 0.9 &  1.1 & 1.2 & 1.7 & 1.2 &  0.9& 2.0 & 1.4\\
MetaWN& 1.9 & 1.0 & 3.9  & 2.0 & 1.5 & 1.0 & 0.9 & 23.0 & 4.4\\\hline

\multicolumn{10}{c}{RoBERTa-large} \\
    \rowcolor{mygray}
Full Clean& 0.07 & 0.33 & 0.16 & 0.59 & 0.7&0.7&0.8&2.0 & 0.7\\
C&  1.8& 9.6 & 7.8 & 1.0 & 1.5&1.2&0.7&4.7  &3.5 \\
W& 0.8 & 0.7 & 0.5 & 2.7 &  2.0&4.1&1.8&2.9  & 1.9\\
C+W& 1.2 & 1.6 & 1.6 & 2.6 & 1.9&1.3&0.7&5.0 & 2.0\\
Snorkel&  0.8& 2.5 & 2.5 & 1.9 & 1.2&2.9&1.9&4.4  & 2.3\\
C+Snorkel& 2.1 & 2.5 & 1.5 & 2.3 & 2.5&3.1&0.7&2.8  & 2.2\\
GLC& 1.7 & 1.0 &  2.1& 1.2 & 28.4 & 1.2 & 0.8 & 3.8 & 5.0\\
MetaWN& 2.0 & 16.6 & 3.0 & 15.6 & 1.4 & 1.5 & 0.6 & 2.9 & 5.5\\
\bottomrule
    \end{tabular}}
    \label{tab:std-results}
\end{table*}

%
%

%

%
%

%
%

%

\paragraph{Analysis of individual weak rules.}
Tables~\ref{tab:rule-results-agnews}-\ref{tab:rule-results-laptopreview} show performance results for each weak rule for the datasets in \name.
We evaluate two different strategies for majority voting in case of an instance that is not covered by any rules:  (1) ``Strict'' counts the instance as misclassified and (2) ``Loose'' assigns a random label to the instance. %
Most rules have very low F1 score while there are a few rules with a relatively high F1 score. 

Figure~\ref{fig:rule-precision-recall-plots} shows the precision-recall scatter plots for each weak rule individually. (We skip the scatter plot for GossipCop as it has just 3 rules.)
Several rules have relatively high precision but most rules have very low recall. 

\begin{table*}[!htb]
    \centering
       \caption{Performance of each rule on AGNews.}
    \resizebox{\linewidth}{!}{
    \begin{tabular}{c|c|c|c||c|c|c||c|c|c||c|c|c}
    \toprule
      \multicolumn{13}{c}{\textbf{AG News}} \\
     	& \multicolumn{3}{c||}{\textbf{unlabeled}}
     	&	\multicolumn{3}{c||}{\textbf{train}} & \multicolumn{3}{c||}{\textbf{validation}} & \multicolumn{3}{c}{\textbf{test}} \\
     \textbf{Rule} & \textbf{Prec} & \textbf{Rec} & \textbf{F1} & \textbf{Prec} & \textbf{Rec} & \textbf{F1} & \textbf{Prec} & \textbf{Rec} & \textbf{F1} & \textbf{Prec} & \textbf{Rec} & \textbf{F1}\\
    \midrule   
    rule 1& 0.179&0.078&0.109&0.179&0.114&0.140&0.182&0.077&0.108&0.180&0.079&0.110\\
    rule 2& 0.157&0.082&0.108&0.156&0.121&0.136&0.159&0.082&0.108&0.154&0.081&0.106\\
    rule 3& 0.162&0.093&0.118&0.162&0.134&0.147&0.160&0.094&0.118&0.166&0.094&0.120\\
    rule 4& 0.192&0.011&0.021&0.192&0.018&0.033&0.192&0.012&0.022&0.193&0.011&0.021\\
    rule 5& 0.187&0.064&0.095&0.189&0.090&0.122&0.190&0.067&0.099&0.188&0.068&0.099\\
    rule 6& 0.140&0.053&0.077&0.141&0.100&0.117&0.137&0.054&0.077&0.141&0.051&0.075\\
    rule 7& 0.161&0.052&0.079&0.163&0.096&0.121&0.163&0.050&0.077&0.163&0.051&0.078\\
    rule 8& 0.136&0.114&0.124&0.138&0.168&0.152&0.134&0.113&0.123&0.137&0.118&0.127\\
    rule 9& 0.152&0.007&0.014&0.153&0.011&0.020&0.149&0.007&0.013&0.154&0.008&0.014\\\hline 
    Majority (strict)& 0.649&0.426&0.512&0.814&0.812&0.812&0.645&0.424&0.509&0.650&0.429&\textbf{0.514}\\
    Majority (loose)& 0.618&0.620&0.617&0.814&0.812&0.812&0.611&0.613&0.610&0.618&0.620&0.618\\ %

    \bottomrule
    \end{tabular}}
    \label{tab:rule-results-agnews}
\end{table*}

\begin{table*}[!htb]
    \centering
       \caption{Performance of each rule on IMDB.}
    \resizebox{\linewidth}{!}{
    \begin{tabular}{c|c|c|c||c|c|c||c|c|c||c|c|c}
    \toprule
      \multicolumn{13}{c}{\textbf{IMDB}} \\
     	& \multicolumn{3}{c||}{\textbf{unlabeled}}
     	&	\multicolumn{3}{c||}{\textbf{train}} & \multicolumn{3}{c||}{\textbf{validation}} & \multicolumn{3}{c}{\textbf{test}} \\
     \textbf{Rule} & \textbf{Prec} & \textbf{Rec} & \textbf{F1} & \textbf{Prec} & \textbf{Rec} & \textbf{F1} & \textbf{Prec} & \textbf{Rec} & \textbf{F1} & \textbf{Prec} & \textbf{Rec} & \textbf{F1}\\
    \midrule   
rule 1& 0.182&0.001&0.001&0.000&0.000&0.000&0.333&0.000&0.001&0.000&0.000&0.000\\
rule 2& 0.000&0.000&0.000&0.000&0.000&0.000&0.000&0.000&0.000&0.000&0.000&0.000\\
rule 3& 0.000&0.000&0.000&0.000&0.000&0.000&0.000&0.000&0.000&0.000&0.000&0.000\\
rule 4& 0.497&0.405&0.446&0.502&0.478&0.489&0.505&0.404&0.448&0.513&0.423&\textbf{0.463}\\
rule 5& 0.538&0.044&0.073&0.549&0.045&0.075&0.408&0.039&0.067&0.481&0.046&0.077\\
rule 6& 0.000&0.000&0.000&0.000&0.000&0.000&0.000&0.000&0.000&0.000&0.000&0.000\\
rule 7& 0.457&0.109&0.176&0.463&0.120&0.190&0.448&0.095&0.156&0.459&0.115&0.183\\
rule 8& 0.655&0.006&0.012&0.644&0.004&0.009&0.630&0.008&0.015&0.667&0.008&0.015\\\hline 
Majority (strict)& 0.495&0.426&0.457&0.749&0.745&0.745&0.501&0.423&0.458&0.511&0.448&\textbf{0.476}\\
Majority (loose)& 0.708&0.707&0.706&0.749&0.745&0.745&0.710&0.708&0.708&0.740&0.739&0.739\\

    \bottomrule
    \end{tabular}}
    \label{tab:rule-results-imdb}
\end{table*}

\begin{table*}[!htb]
    \centering
       \caption{Performance of each rule on Yelp.}
    \resizebox{\linewidth}{!}{
    \begin{tabular}{c|c|c|c||c|c|c||c|c|c||c|c|c}
    \toprule
      \multicolumn{13}{c}{\textbf{Yelp}} \\
     	& \multicolumn{3}{c||}{\textbf{unlabeled}}
     	&	\multicolumn{3}{c||}{\textbf{train}} & \multicolumn{3}{c||}{\textbf{validation}} & \multicolumn{3}{c}{\textbf{test}} \\
     \textbf{Rule} & \textbf{Prec} & \textbf{Rec} & \textbf{F1} & \textbf{Prec} & \textbf{Rec} & \textbf{F1} & \textbf{Prec} & \textbf{Rec} & \textbf{F1} & \textbf{Prec} & \textbf{Rec} & \textbf{F1}\\
    \midrule   
rule 1& 0.642&0.047&0.085&0.638&0.053&0.094&0.643&0.052&0.093&0.614&0.042&0.076\\
rule 2& 0.214&0.029&0.051&0.221&0.036&0.063&0.213&0.028&0.050&0.239&0.031&0.054\\
rule 3& 0.501&0.328&0.371&0.504&0.393&0.419&0.514&0.338&0.381&0.492&0.324&\textbf{0.367}\\
rule 4& 0.498&0.064&0.114&0.485&0.081&0.139&0.501&0.069&0.121&0.491&0.066&0.117\\
rule 5& 0.502&0.101&0.163&0.503&0.122&0.191&0.489&0.090&0.147&0.519&0.105&0.168\\
rule 6& 0.426&0.035&0.065&0.433&0.046&0.083&0.417&0.036&0.066&0.398&0.036&0.066\\
rule 7& 0.486&0.044&0.081&0.484&0.053&0.095&0.509&0.044&0.081&0.479&0.039&0.071\\
rule 8& 0.553&0.049&0.085&0.556&0.060&0.103&0.515&0.049&0.086&0.553&0.053&0.092\\\hline
Majority (strict)& 0.508&0.389&0.411&0.762&0.700&0.692&0.515&0.392&0.415&0.498&0.381&\textbf{0.404}\\
Majority (loose)& 0.710&0.677&0.663&0.762&0.700&0.692&0.719&0.683&0.671&0.706&0.672&0.659\\

    \bottomrule
    \end{tabular}}
    \label{tab:rule-results-yelp}
\end{table*}

\begin{table*}[!htb]
    \centering
       \caption{Performance of each rule on GossipCop.}
    \begin{tabular}{c|c|c|c||c|c|c||c|c|c}
    \toprule
      \multicolumn{10}{c}{\textbf{GossipCop}} \\
     	&	\multicolumn{3}{c||}{\textbf{train}} & \multicolumn{3}{c||}{\textbf{validation}} & \multicolumn{3}{c}{\textbf{test}} \\
     \textbf{Rule} & \textbf{Prec} & \textbf{Rec} & \textbf{F1} & \textbf{Prec} & \textbf{Rec} & \textbf{F1} & \textbf{Prec} & \textbf{Rec} & \textbf{F1}\\
    \midrule   
rule 1&0.632&0.629&0.627 &0.614&0.610&0.607 &0.629&0.627&0.625 \\
rule 2& 0.648&0.622&0.604 &0.643&0.620&0.604 & 0.658&0.630&0.613
  \\
rule 3&0.740&0.731&0.728 &0.754&0.746&0.744 & 0.732&0.726&0.724
\\\hline 
majority &0.758&0.732&0.725 &0.757&0.728&0.721& 0.760&0.740& \textbf{0.735}
\\
    \bottomrule
    \end{tabular}%
    \label{tab:rule-results-GossipCop}
\end{table*}

\begin{table*}[!htb]
    \centering
       \caption{Performance of each rule on NCBI.}
    \begin{tabular}{c|c|c|c||c|c|c||c|c|c}
    \toprule
      \multicolumn{10}{c}{\textbf{NCBI}} \\
     	&	\multicolumn{3}{c||}{\textbf{train}} & \multicolumn{3}{c||}{\textbf{validation}} & \multicolumn{3}{c}{\textbf{test}} \\
     \textbf{Rule} & \textbf{Prec} & \textbf{Rec} & \textbf{F1} & \textbf{Prec} & \textbf{Rec} & \textbf{F1} & \textbf{Prec} & \textbf{Rec} & \textbf{F1}\\
    \midrule   
rule 1& 0.490 & 0.025 & 0.047& 0.460 & 0.066 & 0.116& 0.537 & 0.031 & 0.058 \\
rule 2& 0.514 & 0.017 & 0.034& 0.140 & 0.010 & 0.019& 0.349 & 0.016 & 0.030 \\
rule 3& 0.317 & 0.035 & 0.064& 0.241 & 0.018 & 0.033& 0.295 & 0.024 & 0.045 \\
rule 4& 0.875 & 0.219 & 0.350& 0.911 & 0.118 & 0.208& 0.807 & 0.172 & \textbf{0.283} \\
rule 5& 0.823 & 0.412 & 0.549& 0.707 & 0.445 & 0.546& 0.793 & 0.412 & \textbf{0.542} \\
rule 6& 0.678 & 0.037 & 0.071& 0.794 & 0.035 & 0.066& 0.667 & 0.030 & 0.057 \\
rule 7& 0.227 & 0.002 & 0.004& 0.333 & 0.001 & 0.003& 0.000 & 0.000 & 0.000 \\
rule 8& 0.250 & 0.001 & 0.001& 0.000 & 0.000 & 0.000& 0.000 & 0.000 & 0.000 \\
rule 9& 0.000 & 0.000 & 0.000& 0.000 & 0.000 & 0.000& 0.000 & 0.000 & 0.000 \\
rule 10& 0.000 & 0.000 & 0.000& 0.000 & 0.000 & 0.000& 0.000 & 0.000 & 0.000 \\
rule 11& 0.000 & 0.000 & 0.000& 0.000 & 0.000 & 0.000& 0.000 & 0.000 & 0.000 \\
rule 12& 0.325 & 0.016 & 0.031& 0.036 & 0.001 & 0.002& 0.375 & 0.013 & 0.024 \\\hline 
Majority& 0.749 & 0.637 & 0.688 & 0.659 & 0.566 & 0.609 & 0.716 & 0.590 & \textbf{0.647}\\
    \bottomrule
    \end{tabular}%
    \label{tab:rule-results-NCBI}
\end{table*}

\begin{table*}[!htb]
    \centering
       \caption{Performance of each rule on WikiGold.}
    \begin{tabular}{c|c|c|c||c|c|c||c|c|c}
    \toprule
      \multicolumn{10}{c}{\textbf{WikiGold}} \\
     	&	\multicolumn{3}{c||}{\textbf{train}} & \multicolumn{3}{c||}{\textbf{validation}} & \multicolumn{3}{c}{\textbf{test}} \\
     \textbf{Rule} & \textbf{Prec} & \textbf{Rec} & \textbf{F1} & \textbf{Prec} & \textbf{Rec} & \textbf{F1} & \textbf{Prec} & \textbf{Rec} & \textbf{F1}\\
    \midrule   
rule 0 & 0.590 & 0.406 & 0.481  & 0.539 & 0.399 & 0.459  & 0.591 & 0.397 & 0.475 \\                                         
rule 1 & 0.593 & 0.538 & 0.564  & 0.597 & 0.558 & 0.577  & 0.624 & 0.557 & 0.589 \\
                  
rule 2 & 0.252 & 0.059 & 0.095  & 0.235 & 0.049 & 0.081  & 0.229 & 0.059 & 0.093 \\

rule 3 & 0.226 & 0.060 & 0.095  & 0.193 & 0.049 & 0.078  & 0.211 & 0.061 & 0.095 \\

rule 4 & 0.621 & 0.091 & 0.158  & 0.596 & 0.086 & 0.150  & 0.633 & 0.101 & 0.175 \\

rule 5 & 0.776 & 0.137 & 0.233  & 0.814 & 0.147 & 0.249  & 0.886 & 0.104 & 0.186 \\

rule 6 & 0.773 & 0.137 & 0.233  & 0.814 & 0.147 & 0.249  & 0.886 & 0.104 & 0.186 \\

rule 7 & 0.576 & 0.092 & 0.159  & 0.542 & 0.080 & 0.139  & 0.587 & 0.099 & 0.169 \\

rule 8 & 0.558 & 0.030 & 0.058  & 0.471 & 0.025 & 0.047  & 0.684 & 0.035 & 0.066 \\

rule 9 & 0.547 & 0.030 & 0.058  & 0.471 & 0.025 & 0.047  & 0.684 & 0.035 & 0.066 \\

rule 10 & 0.875 & 0.020 & 0.038  & 0.857 & 0.037 & 0.071 & 1.000 & 0.024 & 0.047 \\

rule 11 & 0.862 & 0.020 & 0.038  & 0.857 & 0.037 & 0.071  & 1.000 & 0.024 & 0.047 \\

rule 12 & 0.885 & 0.177 & 0.295  & 0.864 & 0.215 & 0.344  & 0.857 & 0.176 & 0.292 \\

rule 13 & 0.869 & 0.178 & 0.296  & 0.855 & 0.218 & 0.347  & 0.825 & 0.176 & 0.290 \\

rule 14 & 0.780 & 0.352 & 0.485  & 0.803 & 0.387 & 0.522  & 0.781 & 0.315 & 0.449 \\

rule 15 & 0.758 & 0.353 & 0.482 & 0.768 & 0.387 & 0.514 & 0.727 & 0.312 & 0.437 \\\hline 

Majority & 0.490 & 0.564 & 0.524 & 0.488 & 0.558 & 0.521 & 0.490 & 0.560 & 0.522 \\

    \bottomrule
    \end{tabular}%
    \label{tab:rule-results-BC5CDR}
\end{table*}

\begin{table*}[!htb]
    \centering
       \caption{Performance of each rule on LaptopReview.}
    \begin{tabular}{c|c|c|c||c|c|c||c|c|c}
    \toprule
      \multicolumn{10}{c}{\textbf{LaptopReview}} \\
     	&	\multicolumn{3}{c||}{\textbf{train}} & \multicolumn{3}{c||}{\textbf{validation}} & \multicolumn{3}{c}{\textbf{test}} \\
     \textbf{Rule} & \textbf{Prec} & \textbf{Rec} & \textbf{F1} & \textbf{Prec} & \textbf{Rec} & \textbf{F1} & \textbf{Prec} & \textbf{Rec} & \textbf{F1}\\
    \midrule   
rule 1& 0.000 & 0.000 & 0.000& 0.000 & 0.000 & 0.000& 0.000 & 0.000 & 0.000 \\
rule 2& 0.679 & 0.595 & 0.634& 0.656 & 0.584 & 0.618& 0.722 & 0.512 & \textbf{0.599} \\
rule 3& 0.667 & 0.003 & 0.006& 1.000 & 0.004 & 0.008& 0.500 & 0.003 & 0.006 \\
rule 4& 0.500 & 0.006 & 0.012& 0.400 & 0.009 & 0.017& 0.750 & 0.009 & 0.018 \\
rule 5& 0.000 & 0.000 & 0.000& 0.000 & 0.000 & 0.000& 0.000 & 0.000 & 0.000 \\
rule 6& 0.423 & 0.006 & 0.011& 0.467 & 0.015 & 0.029& 0.000 & 0.000 & 0.000 \\
rule 7& 1.000 & 0.001 & 0.002& 0.500 & 0.002 & 0.004& 0.000 & 0.000 & 0.000 \\
rule 8& 0.000 & 0.000 & 0.000& 0.000 & 0.000 & 0.000& 0.000 & 0.000 & 0.000 \\
rule 9& 0.000 & 0.000 & 0.000& 0.000 & 0.000 & 0.000& 0.000 & 0.000 & 0.000 \\
rule 10& 0.333 & 0.001 & 0.001& 1.000 & 0.002 & 0.004& 0.000 & 0.000 & 0.000 \\
rule 11& 0.000 & 0.000 & 0.000& 0.000 & 0.000 & 0.000& 0.000 & 0.000 & 0.000 \\
rule 12& 0.735 & 0.013 & 0.026& 0.800 & 0.009 & 0.017& 0.250 & 0.006 & 0.012 \\\hline 
Majority & 0.671 & 0.609 & 0.638 & 0.644 & 0.599 & 0.621 & 0.706 & 0.521 & \textbf{0.600}\\
    \bottomrule
    \end{tabular}%
    \label{tab:rule-results-laptopreview}
\end{table*}

\end{document}